\documentclass{article}
\usepackage[T1]{fontenc}
\usepackage{microtype,graphicx,booktabs,longtable,array,multirow,xurl,placeins,float}
\floatstyle{ruled}
\newfloat{algorithm}{tbp}{loa}
\floatname{algorithm}{Algorithm}
\usepackage{fancyhdr}
\usepackage{natbib}
\usepackage{iclr2027_conference,times}
\iclrfinalcopy
\usepackage{amsmath,amssymb,amsthm}
\usepackage{xcolor}
\usepackage{hyperref}
\usepackage[capitalize,noabbrev]{cleveref}
\crefname{algorithm}{Algorithm}{Algorithms}
\newtheoremstyle{plaintight}{.5pc}{0pt}{\itshape}{}{\bfseries}{.}{.5em}{}
\newtheoremstyle{deftight}{.5pc}{0pt}{\normalfont}{}{\bfseries}{.}{.5em}{}
\theoremstyle{plaintight}
\newtheorem{theorem}{Theorem}
\newtheorem{proposition}[theorem]{Proposition}

\theoremstyle{deftight}

\newcommand{\R}{\mathbb R}
\newcommand{\E}{\mathbb E}
\newcommand{\I}{\mathrm{I}}

\newcommand{\method}{\textsc{MGPA}}
\newcommand{\sg}{s_Q}
\newcommand{\pseudopar}[1]{\par\vspace{-\parskip}\textbf{#1}}
\DeclareMathOperator{\logit}{logit}
\DeclareMathOperator{\Cov}{Cov}
\DeclareMathOperator{\Var}{Var}
\DeclareMathOperator{\tr}{tr}



\usepackage{enumitem}
\setlist[itemize]{leftmargin=1.2em,topsep=2pt,partopsep=0pt,itemsep=1pt,parsep=0pt}
\setlist[enumerate]{leftmargin=1.6em,topsep=2pt,partopsep=0pt,itemsep=1pt,parsep=0pt}
\usepackage{etoolbox}
\AtBeginEnvironment{table}{\setlength{\abovecaptionskip}{0pt}}

\title{Measurement-Gated Provenance Attenuation\\for Frozen EEG Representations}
\author{Anuar Aimoldin, Yankai Chen, Ayana Mussabayeva, Nurdaulet Akhanov, Xue Liu\\
Mohamed bin Zayed University of Artificial Intelligence (MBZUAI)\\
\texttt{anuar.aimoldin@mbzuai.ac.ae}}
\hypersetup{colorlinks=true,linkcolor=black,citecolor=black,urlcolor=blue!55!black,
pdfauthor={Anuar Aimoldin, Yankai Chen, Ayana Mussabayeva, Nurdaulet Akhanov, Xue Liu},pdftitle={Measurement-Gated Provenance Attenuation for Frozen EEG Representations}}
\begin{document}
\maketitle\vspace{-12pt}
\lhead{Preprint}
\raggedbottom

\begin{abstract}
Frozen EEG representations retain acquisition signatures as well as neural activity. Source predictability alone does not identify what should be removed: it can reflect measurement effects or genuine biological and population differences, which should not be erased. We propose \emph{Measurement-Gated Provenance Attenuation} (MGPA), built on one principle: measurement evidence determines where correction may act, and preserved information determines what it should aim for. Paired measurement contrasts define a gate outside which nothing changes; inside it, the source score is moved to the value the preserved coordinates already predict: for a fixed affine score, this retains the same information as any other target set by those coordinates and needs the least expected squared movement. Closed-form and critic-guided iterative constructions apply it without source identity or encoder retraining. Three studies test the principle at increasing distance from its assumptions. Under controlled reference changes, where the source--task association is known exactly, MGPA brings source to near chance with task performance unchanged, whereas erasing what predicts source (LEACE) lowers frozen-task AUROC from $.753$ to $.656$ while barely touching source; ablations attribute the attenuation to the gate's directions and $2.7\times$ less movement to the conditional target. Across recordings from different devices and electrodes, iterative correction lowers source accessibility while preserving or improving task performance. Finally, one iterative map selected on one task and reused unchanged on existing heads for two others raises their worst-association AUROC (the lowest over device--label association shifts) by $.057$ and $.019$ over LEACE, at a cost to those heads while the training association holds. A reusable correction thus shows its value in how an existing predictor behaves once acquisition cues stop being reliable, not only in what a probe can read.
\end{abstract}

\section{Introduction}
\label{sec:introduction}
EEG foundation models provide reusable representations when labeled data are scarce \citep{wang2024eegpt}, but a frozen encoder also retains its sensitivity to the acquisition pipeline: reference, montage, channel configuration, preprocessing and recording workflow alter the signal it receives \citep{nunez2006,dong2019}. Its representations therefore carry signatures of how the EEG was measured as well as the neural activity. These signatures confound data pooled across sites and, most visibly, become a shortcut \citep{geirhos2020,zech2018}: a downstream head trained under one source--label association can fail when the association changes, even for acquisition systems it has seen.

\begin{figure}[t]
\centering
\includegraphics[width=0.946\linewidth]{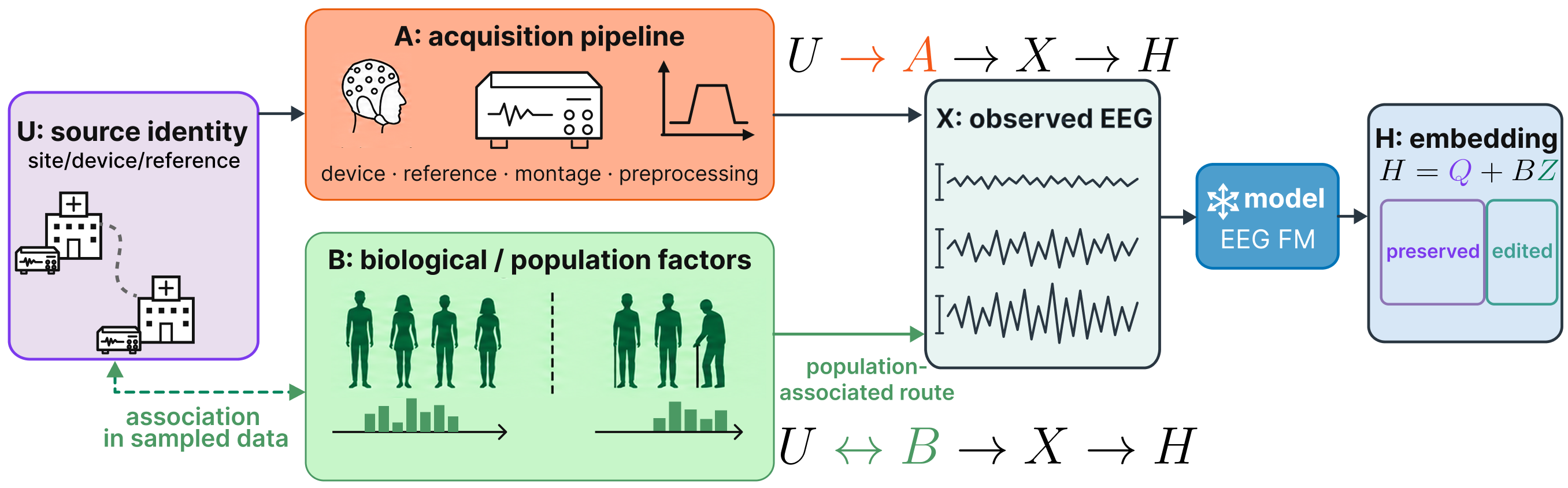}
\caption{\textbf{Two routes to source-predictive frozen EEG representations.}
Source $U$ reaches the representation $H$ of observed EEG $X$ through the
acquisition pipeline ($A$) or through sampling association with biological or
population factors ($B$; route labels local to this diagram). \method\ edits only
the gate coordinates $Z$ of $H=Q+BZ$ ($B$: gate basis, Section~\ref{sec:admissibility}) and preserves $Q$.}
\label{fig:measurement-paths}
\end{figure}

Erasing everything that predicts source is not a safe default, because source is predictable through two routes (Figure~\ref{fig:measurement-paths}): the acquisition pipeline, and the population sampled under each source, which often differs and can matter for the task \citep{yamashita2019}. Correcting only the first route requires two decisions that source labels alone cannot make. \emph{Where to edit:} source-labeled data cannot separate an acquisition effect from a source-associated content difference, but measurement contrasts can, because they compare the same observation under a specified change of the measurement process, such as re-referencing a stored trial or recording one event with simultaneous systems. \emph{What to aim for:} if the preserved coordinates already predict the source, pushing the full source score to a neutral value makes the editable coordinates cancel evidence that is still available, which costs movement without removing more information.

We propose \emph{Measurement-Gated Provenance Attenuation} (\method), a general construction built on one principle: \emph{measurement evidence determines where correction may act, and preserved information determines what it should aim for} (Figure~\ref{fig:theory-geometry}). Paired contrasts define a fixed subspace, the \emph{measurement gate}; every update lies in it, and its complement $Q$ is preserved. Within the gate, the target is the expected source score given $Q$. \emph{Closed-form MGPA} applies one analytic correction along a direction estimated from measurement pairs; \emph{Iterative MGPA} makes bounded updates in the same gate, guided by source predictors (\emph{critics}) refitted after each stage. Our contributions are:
\begin{itemize}
\item \emph{Measurement-gated correction} (Section~\ref{sec:method}). Paired measurement contrasts, not source predictability, decide where a frozen representation may change; two constructions, closed-form and critic-guided iterative, apply this without source identity or encoder retraining, so one stored map serves data whose source is unrecorded or mixed.
\item \emph{A movement-optimal target} (Section~\ref{sec:targets}). For a fixed affine score, every target set by the preserved coordinates retains the same information, and the conditional mean needs the least expected squared movement; in a Gaussian model it beats whole-block replacement.
\item \emph{Evidence} (Section~\ref{sec:experiments}). Controlled SSVEP, where the measurement change is known exactly, isolates each half: the gate beats rank-matched PCA, random and ungated edits at matched movement, and conditional targeting needs $2.7\times$ less movement than unconditional. On recorded device and electrode differences, Iterative MGPA reduces source accessibility, preserving or improving task utility. One P300-selected map reused on frozen N170/MMN heads improves worst-association AUROC over LEACE, at a cost to those heads under the training association.
\end{itemize}
\section{Related Work}
\label{sec:related}
\textbf{Concept erasure and conditional invariance.}
Concept erasure removes whatever predicts a designated attribute while limiting changes to other information. INLP repeatedly removes linearly predictive directions \citep{ravfogel2020}, and LEACE provides a closed-form affine eraser with minimum distortion and linear guardedness \citep{belrose2023}. Nonlinear methods extend the correction geometry: IGBP refits critics after gradient-based projections \citep{iskander2023}, while MANCE constrains edits to local representation-manifold tangents \citep{avitan2026mance}. Other approaches specify what should be preserved. SPLINCE preserves covariance with a supplied task label \citep{holstege2025splince}; CIRCE learns nuisance independence conditional on task information \citep{pogodin2023circe}; and conditional probing measures information beyond a baseline \citep{hewitt2021conditional}. \method\ instead lets measurement contrasts, not source predictability, decide where an edit is permitted, and needs no task label: its target is conditioned on the preserved complement.
\pseudopar{Correction from measurement pairs.}
Traveling-subject designs separate measurement bias from population sampling bias \citep{yamashita2019}. Noisy Counterfactual Matching trains a task predictor invariant to the leading singular directions of paired differences by projecting inputs onto their complement \citep{bai2025ncm}; FEATMAP fits an affine map between paired medical embedding domains without task labels \citep{donle2026featmap}. These methods share \method's evidence but not its action: NCM uses task labels to make a predictor ignore that subspace, and FEATMAP moves every input to a reference domain and needs its source at application. Excluding the whole subspace also drops directions with no conditional source information but possible task information; \method\ edits within the span toward a conditional target and keeps the rest. Post-Training Augmentation Invariance \citep{eikenberry2026} and PISCO \citep{ngweta2023} also adapt frozen representations without retraining, but augmentations or style priors, not calibration, decide what may change.
\pseudopar{Distribution alignment for EEG transfer.}
CORAL matches domain covariances \citep{sun2016coral}, Euclidean alignment normalizes EEG covariance \citep{he2020ea}, and CMMN aligns spectral distributions \citep{gnassounou2023}. Riemannian Procrustes additionally uses a class-informed rotation \citep{rodrigues2019rpa}; for cross-headset classification, AwAR applies adaptation regularization and SDDA uses spatial distillation with distribution alignment \citep{wu2016awar,liu2026sdda}. These methods realign whole distributions and use the source of each input at application. \method\ applies one stored correction to each input without its source label.
\section{Problem Formulation}
\label{sec:admissibility}
\begin{figure}[t]
\centering
\includegraphics[width=\linewidth]{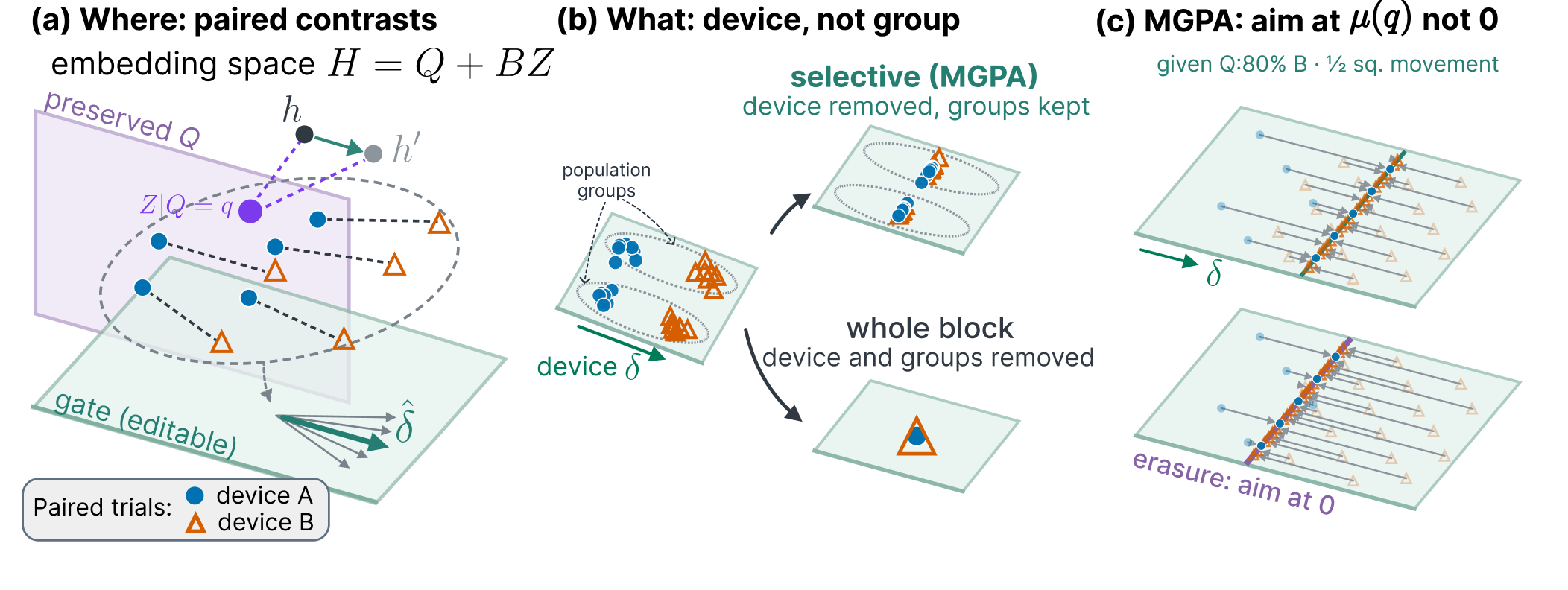}
\caption{\textbf{Three choices in measurement-gated correction.}
(a) Embedding space $H=Q+BZ$. Dashed segments join the two recordings of one
trial; their differences (grey fan) span the gate, and their mean
$\widehat\delta$ is the closed-form direction; a new input $h$ moves to $h'$
along $\widehat\delta$. (b) Slice at fixed $Q=q$; rows are population groups.
Selective correction moves points only along the device direction $\delta$;
whole-block replacement collapses them. (c) Same slice; arrows show the
movement to the zero target of erasure and to $\mu(q)=\E[s\mid Q=q]$, the mean source log-odds score given $q$.
Gaussian illustration (source means $(\pm1,0)$, covariance $I_2$,
$\Pr(U=1\mid q)=.8$), not EEG data.}
\label{fig:theory-geometry}
\end{figure}

\textbf{Representations and calibration.}
Let $H\in\R^d$ denote a representation from a frozen encoder or fixed feature extractor. The source $U\in\{0,1\}$ identifies an acquisition system or constructed measurement view, and $Y$ denotes the downstream prediction target. Calibration provides paired representations $H_i^{(a)},H_i^{(b)}$, where $i$ indexes matched units and $a,b$ identify two acquisition conditions or transformations of the same stored trial. We assume that paired views share a content contribution and have additive measurement responses in a fixed subspace. The contrast $\Delta_i^{ab}=H_i^{(a)}-H_i^{(b)}$ then cancels the shared contribution. Because the remaining response can depend on content in magnitude and direction, calibration contrasts span a subspace rather than a single direction; recovering that span requires sufficiently diverse contrasts \citep{bai2025ncm}.
\pseudopar{Editable and preserved coordinates.}
Let the $r$ orthonormal columns of $B\in\R^{d\times r}$ span the directions supported by calibration, where $r$ is the gate rank. The measurement gate $P=BB^\top$ is the orthogonal projection onto this subspace. We write the editable coordinates as $Z=B^\top H\in\R^r$ and the preserved complement as $Q=q(H)=(I-P)H$, with $I$ the identity. Lowercase $h,z,q$ denote realized values of $H,Z,Q$. A correction map $R$ is admissible if
\begin{equation}
 R(h)=q(h)+B\psi(h),
 \label{eq:admissible}
\end{equation}
where $\psi(h)\in\R^r$ gives the corrected editable coordinates. Since $Q$ is recoverable from every output, the chain rule for mutual information $\I$, conditional on the fitted gate, gives
\begin{equation}
 \I(U;R(H))=\I(U;Q)+\I(U;R(H)\mid Q).
 \label{eq:floor}
\end{equation}
The first term, the source information already in the preserved coordinates, cannot be changed by an admissible correction, whether it stems from source-associated content or from measurement variation the gate missed; balanced source counts do not make it vanish. Only the second term is under the correction's control. Conversely, $Z$ can carry task variation, so permission to edit it is not an instruction to discard it. The problem is to attenuate the second term while moving $Z$ minimally.
\pseudopar{Estimating the measurement gate.}
After featurewise standardization on the fitting data, $B$ holds the leading $r$ right singular vectors of the unit-normalized contrasts $\Delta_i^{ab}$, so recurring directions count independently of their magnitude. The rank $r$ is set by retained energy or development-data selection, and the fitted gate then stays fixed; a zero-rank gate returns the identity map (Appendix~\ref{app:protocols}).
\section{Measurement-Gated Provenance Attenuation}
\label{sec:method}
\label{sec:algorithm}

Given the gate, MGPA chooses a correction target and an update within the editable subspace (Figure~\ref{fig:theory-geometry}). The target must account for source evidence in the preserved coordinates: if $Q$ alone assigns probability $.8$ to one of two equally likely sources, forcing a source predictor on the full representation to output $.5$ asks the editable coordinates to offset evidence that remains available in $Q$, adding movement without removing information. We therefore target the expected source score given $Q$, the \emph{conditional-mean anchor}; this is an expectation of the score, not the source probability given $Q$. For an affine score it follows from $\E[Z\mid Q]$; for a learned critic we regress its scores on $Q$. 

\subsection{Closed-form Correction}
\label{sec:mgpa-cf}
Closed-form MGPA uses measurement pairs to estimate a single correction direction and adjusts the corresponding deviation from the conditional mean. Specifically, we set $\widehat\delta\in\R^r$ to the mean projected calibration contrast $B^\top\Delta_i^{ab}$; the update below is invariant to its sign. On the fitting data, ridge regression estimates the conditional mean $\widehat m_Z(q)\approx\E[Z\mid Q=q]$, and a joint regression of $Z$ on $Q$ and source $U$ provides residuals for a regularized covariance estimate $C\succ0$. For a nonzero fitted direction, the update is
\begin{equation}
 z'=z-\alpha\,\widehat\delta\,
 \frac{\widehat\delta^\top C^{-1}[z-\widehat m_Z(q)]}
 {\widehat\delta^\top C^{-1}\widehat\delta},\qquad q'=q.
 \label{eq:closed-form-mgpa}
\end{equation}
The output is $R(h)=q+Bz'$, and $\alpha$ controls correction strength, with $\alpha=1$ the full step. The fraction is the $C^{-1}$-weighted projection coefficient of the residual $z-\widehat m_Z(q)$ onto $\widehat\delta$. At $\alpha=1$, this is the score-level projection in Proposition~\ref{prop:parallel-levels}, with score coefficient $w=C^{-1}\widehat\delta$, quadratic movement metric $\Omega=C^{-1}$, and anchor $w^\top\widehat m_Z(q)$. The ridge mean estimate makes the complete map affine, so no separate source critic is required. Measurement pairs determine the direction, while the preserved context determines the reference value.

\subsection{Iterative Correction with Source Critics}
\label{sec:mgpa-iter}
The closed-form map acts along a single estimated mean direction. To allow observation-dependent edit directions within the gate, Iterative MGPA uses learned source scores to guide bounded corrections. At stage $k$, $H^{(k)}$ denotes the representation entering that stage, and critic $j=1,\ldots,J$ outputs a source-logit difference $s_j^{(k)}(h)$. On the fitting split (FIT), we regress the current scores on the original complement to estimate the anchor $a_j^{(k)}$ and define the score residual $e_j^{(k)}$:
\begin{equation}
 a_j^{(k)}(q)\approx\E[s_j^{(k)}(H^{(k)})\mid Q=q],\qquad
 e_j^{(k)}(h)=s_j^{(k)}(h)-a_j^{(k)}(q(h)).
 \label{eq:targets}
\end{equation}
Because $q(h+B\Delta z)=q(h)$ for any editable update $\Delta z\in\R^r$, the anchor stays constant during that update. Suppressing the stage index, we stack the critics' residuals in $e\in\R^J$ and form the gated Jacobian $A\in\R^{J\times r}$ with rows $A_j=\nabla_hs_j(h)^\top B$, so that $e(h+B\Delta z)\approx e(h)+A\Delta z$. A positive diagonal weighting $W$, based on gated gradient norms with a positive stabilizer, balances the critic scales; with $\bar A=WA$ and $\bar e=We$, we solve
\begin{equation}
 \Delta z_* = \arg\min_{\|\Delta z\|\le\rho}
 \left\{\tfrac12\|\bar A\Delta z+\bar e\|^2
 +\tfrac\tau2\|\Delta z\|^2\right\},
 \qquad h^+=h+B\Delta z_*,\quad\tau>0.
 \label{eq:tr-problem}
\end{equation}
The first term reduces the linearized score residuals, the second penalizes movement, and the radius $\rho>0$ bounds the displacement at each stage. Linear critics supply fixed gradients, nonlinear critics observation-dependent ones. After updating FIT representations, we refit the critics and their anchors to address the remaining source structure. Each stage preserves $Q$, so their composition does as well in exact arithmetic. Appendix~\ref{app:solver} gives the weights and trust-region solver.

\subsection{Fitting and Reuse}
At fixed settings, both adapters are fitted with measurement calibration and source labels, without downstream task labels. Deployment applies the fitted affine map or replays the stored critics and anchors, requiring neither source identity nor encoder updates. Correction strength is set by $\alpha$ for the closed-form map or by an iterative prefix, optionally with a fractional final stage; a closed-form step with $\alpha\neq1$ is invertible given $Q$ and therefore preserves exact conditional source information (Appendix~\ref{app:partial-step}), yet can still change the predictions of a fixed head. The operating point can be fixed in advance, matched by validation movement, or selected with development task labels; in the reuse study, P300 labels select it and the map is then held fixed on the target tasks. Algorithm~\ref{fig:algorithm-main} and Appendix~\ref{app:protocols} give the fitting procedure and numerical settings.
\section{Theoretical Analysis}
\label{sec:targets}
Why anchor to the conditional mean rather than to zero? The Bayes source score, the posterior log-odds of source $1$ versus source $0$, splits into the log-odds carried by $Q$, $\sg(q)=\logit p(q)$ with $p(q)=\Pr(U=1\mid Q=q)\in(0,1)$, which no admissible correction can change, and the log-likelihood ratio of the editable coordinates given $Q$ (Appendix~\ref{app:bayes-split}). A zero target asks $Z$ to cancel $\sg$; the conditional-mean target does not. Proposition~\ref{prop:parallel-levels} shows that, for a fixed affine score and metric, all $Q$-determined targets retain the same information and the conditional mean needs the least expected squared movement. In an equal-covariance Gaussian model, Theorem~\ref{thm:compensation} solves the problem of Section~\ref{sec:admissibility} among affine maps: the same map erases conditional source information with the least expected squared movement, zero targeting pays an explicit excess, and selective anchoring is cheaper than replacing the whole editable block. These results explain the closed-form construction and motivate the iterative target; they do not establish optimality for refitted nonlinear trajectories. Theorem~\ref{thm:compensation} assumes the true shift $\delta$; with an estimated $\widehat\delta$, Proposition~\ref{prop:parallel-levels} still applies, but conditional erasure need not hold.

\subsection{Target Choice and Correction Cost}
\label{sec:affine}
To study target choice without requiring a Bayes-optimal predictor, fix a
score $s(q,z)=b(q)+w(q)^\top z$, affine in the editable coordinates, and a
positive-definite correction metric $\Omega(q)$. We condition on the fitted
gate and use nonredundant coordinates for $Q$. The functions $b,w,\Omega$
may depend on $q$, but not on $z$, and $w(q)\ne0$. Correction cost is measured by $\|\Delta z\|_{\Omega(q)}^2=\Delta z^\top\Omega(q)\Delta z$.
For a finite measurable target $a(q)$, we construct the minimum-cost map $T_a$ that reaches that score while preserving $q$. Suppressing the dependence on $q$, define
\begin{equation}
 d_\Omega=w^\top\Omega^{-1}w,\qquad v=\Omega^{-1}w/d_\Omega,\qquad
 T_a(q,z)=\bigl(q,z+v\{a(q)-s(q,z)\}\bigr).
 \label{eq:affine-map}
\end{equation}
Here $v$ is the correction direction scaled to produce a unit change in score, so $T_a$ projects the editable coordinates onto the selected score level. Below, $\sigma(\cdot)$ denotes the generated sigma-algebra, and $N$ is the component of $Z$ that the projection leaves unchanged ($w^\top N=0$). Combining metric projection with the conditional squared-error decomposition yields the following result.

\begin{proposition}[Target choice changes distortion, not retained information]
\label{prop:parallel-levels}
$T_a$ is the unique minimum-$\Omega$-norm edit reaching score $a(q)$.
For any joint law, all such targets generate the same information:
$\sigma(T_a(Q,Z))=\sigma(Q,N)$, where $N=(I-vw^\top)Z$.
Let $\mathcal C_a(q)$ be the expected squared $\Omega(q)$-norm of the
editable displacement given $Q=q$. With finite conditional score second moments,
\begin{equation}
 \mathcal C_a(q)
 =\frac{\Var(s\mid Q=q)+\{\E[s\mid Q=q]-a(q)\}^2}{d_\Omega(q)}.
 \label{eq:affine-cost}
\end{equation}
Consequently, $\mu(q)=\E[s\mid Q=q]$ uniquely minimizes expected squared
movement in the chosen metric within this fixed-score family.
\end{proposition}

Proposition~\ref{prop:parallel-levels} separates coordinate placement from retained information: the corrected editable block is $N+v(a-b)$, so targets differ only by a translation recoverable from $Q$, and all targets retain the same information about any downstream variable, although an unchanged head may respond differently to the coordinates. The conditional mean therefore minimizes movement within this family; the proposition does not assert that the family preserves all task information in the original representation. The argument needs neither Gaussian data nor a Bayes-optimal score. With Euclidean movement, as penalized in the iterative update, $\Omega=I$ and $d_\Omega=\|w\|^2$, giving the same conditional-mean optimum (Appendix~\ref{app:euclidean-anchor}). Conditional source erasure holds precisely when $N\perp U\mid Q$ (Appendix~\ref{app:affine-levels}), which the Gaussian model below makes explicit.

\subsection{Conditional Erasure in a Gaussian Model}
\label{sec:gaussian}
Let $m(q)$ be the midpoint of the two source-conditional means, $\delta$ their difference (source $1$ minus source $0$) and $\Sigma$ their common within-source covariance, with $\delta$ and $\Sigma$ shared across $q$:
\begin{equation}
 Z\mid Q=q,U=u\sim\mathcal N(m(q)+(2u-1)\delta/2,\Sigma),\qquad
 \Sigma\succ0,\quad D=\delta^\top \Sigma^{-1}\delta>0,
 \label{eq:gaussian}
\end{equation}
where $D$ is the squared Mahalanobis separation of the source means. The Bayes score is $s=\sg+\ell$ with $\ell=\delta^\top \Sigma^{-1}(z-m(q))$, and $\E[\ell\mid Q=q]=(2p-1)D/2$, which is nonzero unless $p(q)=1/2$. Proposition~\ref{prop:parallel-levels} with $w=\Sigma^{-1}\delta$ and $\Omega=\Sigma^{-1}$ gives $v=\delta/D$. For a residual target $\kappa(q)=a(q)-\sg(q)$, the resulting map $R_\kappa(q,z)=\bigl(q,\,z+\tfrac{\delta}{D}\{\kappa(q)-\ell(q,z)\}\bigr)$ brings $\ell$ to $\kappa(q)$; $\mathcal C_\kappa(q)$ denotes its expected squared displacement in the $\Sigma^{-1}$ metric, and zero-score targeting uses $\kappa=-\sg$. At each fixed $q$, $R_\kappa$ specializes the binary LEACE geometry \citep{belrose2023}.

\begin{theorem}[Gaussian conditional erasure and target-dependent cost]
\label{thm:compensation}
Under \eqref{eq:gaussian}, every $R_\kappa$ preserves $Q$ and satisfies
$R_\kappa(Q,Z)\perp U\mid Q$, with
\begin{equation}
 \mathcal C_\kappa(q)=1+Dp(1-p)
       +\frac{\{\kappa(q)-(2p-1)D/2\}^2}{D},\qquad p=p(q).
 \label{eq:gauge-cost}
\end{equation}
The cost is minimized at $\kappa_*(q)=\E[\ell\mid Q=q]$, i.e.\ at the total-score target $a_*(q)=\E[s\mid Q=q]$; zero-score targeting incurs the excess
\begin{equation}
 \mathcal C_{-\sg}(q)-\mathcal C_{\kappa_*}(q)
 =\frac{\{\sg(q)+(2p-1)D/2\}^2}{D}\ge0,
 \label{eq:compensation}
\end{equation}
which vanishes only at $p(q)=1/2$. Among $Q$-preserving deterministic maps whose editable output is affine in $Z$ at each fixed $q$, mean anchoring minimizes expected squared movement in the $\Sigma^{-1}$ metric subject to conditional source erasure.
\end{theorem}

Selectivity enters through the cost: replacing the whole editable block by $\E[Z\mid Q]$ costs $r+Dp(1-p)$, whereas selective anchoring costs $1+Dp(1-p)$ and leaves untouched $r-1$ conditionally source-independent directions, which may carry task information (Appendix~\ref{app:completion}). Estimating the anchor adds the cost quantified in Appendix~\ref{app:anchor-estimation}.
\section{Experiments}
\label{sec:experiments}
Three studies test the principle at increasing distance from its assumptions: a controlled setting that isolates the gate and the target (Section~\ref{sec:gate-result}), recordings from different devices and electrodes (Section~\ref{sec:real-results}), and an existing head under an association shift (Section~\ref{sec:recovery}). The recorded studies use the same participants on both sources, so in them source is an acquisition effect by design; the controlled setting adds a source--task association with known ground truth.

\subsection{Experimental setup}
\label{sec:protocol}
\textbf{Metrics.}
Source accessibility is the maximum AUROC, $S$, over a bank of
independently trained source readers; it measures what can be read, not
what a task head uses. Exact conditional erasure (Section~\ref{sec:targets}), accessibility to independent readers and the behavior of an existing head are three distinct objects; the metrics report the latter two. Task AUROC uses heads trained before correction
($T_f$, frozen) or after it ($T_r$, refitted), separating compatibility
with an existing predictor from utility after retraining.  Movement $M$ is the participant-mean relative
embedding displacement. The reuse study instead evaluates fixed heads
across device--label associations, alongside a control head trained without that association. $\downarrow$/$\uparrow$ mark quantities to minimize/maximize.
\pseudopar{Fitting and comparisons.}
Participants have disjoint roles for adapter fitting (FIT), validation
(VAL), reader and head training (HEAD) and evaluation (EVAL); in reuse, a
development set (DEV) replaces VAL and selects correction strength using
P300 labels, after which the map is fixed on all evaluation tasks. MGPA
fitting uses source labels and calibration contrasts, never downstream
labels. Comparisons use each method's complete fitted map (native output) unless validation movement is matched (Appendix~\ref{app:protocols}). Identity, LEACE and IGBP are source-blind; CORAL and paired FEATMAP need source identity at application. Iterative estimates average three fits; brackets give paired 95\% participant-bootstrap intervals conditional on fitted maps and readers (Appendix~\ref{app:protocols}).

\subsection{Controlled SSVEP: correction, gate and target}
\label{sec:gate-result}
We use wet trials from the 102-participant SSVEP collection
\citep{zhu2021ssvep}, encoding each trial under common-average and Oz
references. Full reference contrasts provide calibration; evaluated views
use the same midpoint with the contrast scaled by $\gamma=.005$. Source assignment is
associated with stimulus frequency during fitting and balanced during
evaluation. This emulates the second route in Figure~\ref{fig:measurement-paths}: source prediction also reflects task information.
\begin{figure}[t]
\centering
\includegraphics[width=\linewidth]{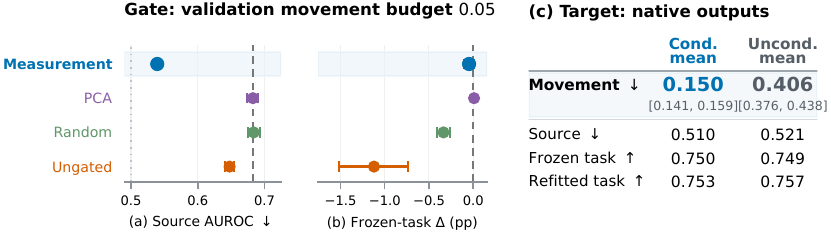}
\caption{\textbf{Effects of the gate and target in Iterative MGPA.}
\emph{(a--b)} Source AUROC and paired frozen-task change from identity (pp: percentage points)
at a common $.05$ validation movement budget; dashed: identity, dotted:
chance; random averages five orientations. \emph{(c)} Conditional versus
empirical unconditional mean-score anchoring at native 48-stage outputs.
Three fits, 81 participants; whiskers and movement brackets give 95\%
participant-bootstrap intervals.}
\label{fig:gate}
\label{fig:target-cost}
\end{figure}

\pseudopar{Selective correction.}
Closed-form MGPA brings source accessibility near chance while retaining frozen-task performance, whereas LEACE lowers frozen-task AUROC to $.656$ but barely lowers source ($.683$ to $.660$; Table~\ref{tab:controlled-cf}). Relative to LEACE,
the paired source change is $-.1496$ $[-.1580,-.1426]$ and the frozen-task
gain is $.0964$ $[.0831,.1097]$. CORAL and FEATMAP achieve this with less movement but route each input by its known measurement condition; MGPA uses one source-blind map.

\begin{table}[t]
\centering\small
\caption{\textbf{Selective correction on controlled SSVEP.} Native outputs;
MGPA-CF: Closed-form MGPA; ref.~0/1: target view (scaled common-average/Oz).
One reader realization over four rotations, 81 evaluation participants;
paired contrasts in Appendix~\ref{app:closed-form-controlled}.}
\label{tab:controlled-cf}
\begingroup\small\setlength{\tabcolsep}{4pt}
\begin{tabular}{@{}llrrrr@{}}
\toprule
Interface & Method & Source $\downarrow$ & Frozen $\uparrow$ & Refitted $\uparrow$ & Movement \\
\midrule
\multirow{3}{*}{\shortstack[l]{\emph{Source-blind}\\\emph{correction}}} & Identity & 0.683 & 0.753 & 0.753 & 0.000 \\
 & LEACE & 0.660 & 0.656 & 0.665 & 0.120 \\
 & \textbf{MGPA-CF} & 0.510 & 0.753 & 0.753 & 0.028 \\
\midrule
\multirow{4}{*}{\shortstack[l]{\emph{Source-aware}\\\emph{alignment}}} & CORAL $\to$ ref. 0 & 0.502 & 0.753 & 0.753 & 0.012 \\
 & CORAL $\to$ ref. 1 & 0.502 & 0.753 & 0.753 & 0.012 \\
 & FEATMAP $\to$ ref. 0 & 0.502 & 0.753 & 0.753 & 0.012 \\
 & FEATMAP $\to$ ref. 1 & 0.502 & 0.753 & 0.753 & 0.012 \\
\bottomrule
\end{tabular}\endgroup

\end{table}

\textbf{Constraining critic-driven corrections (Iterative MGPA).}
Do critic-driven updates benefit from measurement-derived rather than generic edit permissions? The arms differ only in the permitted subspace and share critics, solver, stage budget and the original complement as anchor input, which only the measurement gate preserves. At a common validation movement budget, the measurement gate attenuates source more strongly than same-rank PCA, random and ungated controls (Figure~\ref{fig:gate}a--b); its frozen-task change from identity is $-.0005$ $[-.0010,.0001]$ and its frozen-task advantage over ungated correction is $.0107$ $[.0069,.0147]$. The gain therefore depends on which directions calibration identifies, not on restricting rank or movement alone (Appendix~\ref{app:gate-attribution}).

\textbf{Conditional targets.}
\label{sec:ablations}%
Conditional-mean anchoring reduces movement and source accessibility, at a small cost in refitted-task performance. In Figure~\ref{fig:target-cost}c, it uses
$M=.1500$, whereas the empirical unconditional mean-score target needs
$.4063$, about 2.7 times more. Source AUROC is lower
($.5105$ versus $.5214$), with no clear frozen-task difference, whereas
refitted-task AUROC favors the unconditional target ($.7574$ versus
$.7527$). These learned trajectories refit their critics separately, so they support a correction-cost advantage rather than test the fixed-score setting of Section~\ref{sec:targets}, in which all targets retain the same information. Target definitions and paired contrasts: Appendix~\ref{app:target-controls}.

\subsection{Source attenuation and task utility on recorded EEG}
\label{sec:real-results}
Iterative MGPA attenuates source on both recorded datasets and Closed-form MGPA on SSVEP, with task AUROC unchanged or improved; the nonlinear critic helps on temporal N170 features, with no measurable advantage on frozen EEGPT SSVEP (Table~\ref{tab:real-main}).  N170
face/watch discrimination uses 240-dimensional temporal features from
simultaneous Neuroscan/Flex recordings \citep{williams2020flex}, i.e.\
physiologically paired trials; wet/dry SSVEP uses frozen EEGPT
representations, and because its sessions have no trial correspondence,
calibration uses six FIT-local spectral and spatial transformations of each
trial. SSVEP thus tests whether the corrections still attenuate source when physiological pairing is unavailable.
\pseudopar{Temporal N170.}
Adding the nonlinear critic improves source attenuation and refitted-task
performance on N170. Relative to the linear-only ablation, the paired
source AUROC change is $-.0413$ $[-.0647,-.0210]$ and the refitted-task gain
is $.0171$ $[.0005,.0384]$, at nearly identical movement ($.3469$ versus
$.3449$; Table~\ref{tab:real-main}A). Full Iterative
MGPA also improves frozen and refitted task AUROC over identity, although
the refitted gain is concentrated in Neuroscan-to-Flex transfer
(Appendix~\ref{app:retained-results}).

Iterative MGPA also improves N170 task utility relative to IGBP
\citep{iskander2023}. Its native refitted-task gain is $.0361$
$[.0306,.0405]$, with less movement ($.3469$ versus $.4973$). At a common
$.25$ validation movement budget, it achieves lower source AUROC and
higher frozen/refitted task AUROC
(Appendix~\ref{app:igbp-comparison}). Source-aware FEATMAP toward Neuroscan
attains stronger attenuation and higher task scores at a larger
displacement.

\begin{table}[t]
\centering\small
\caption{\textbf{Source attenuation and task utility on recorded EEG.}
Native outputs (movements differ) for N170 (A, 10 participants) and wet/dry
SSVEP (B, 21). $S$: maximum source AUROC over the eleven-reader bank (A) or
the 41-reader full-input bank (B); $T_f/T_r$: frozen/refitted task AUROC;
$M$: participant-mean relative movement; MGPA-Iter: Iterative MGPA.
Ref.~0/1: Neuroscan/Flex (A), dry/wet (B). N/A: FEATMAP needs paired trials,
unavailable in B.}
\label{tab:real-main}
\begingroup\small\setlength{\tabcolsep}{3pt}
\begin{tabular*}{\textwidth}{@{\extracolsep{\fill}}llrrrrrrrr@{}}
\toprule
 & & \multicolumn{4}{c}{A. Temporal N170} & \multicolumn{4}{c}{B. Recorded SSVEP} \\
\cmidrule(lr){3-6}\cmidrule(lr){7-10}
Interface & Method & $S\downarrow$ & $T_f\uparrow$ & $T_r\uparrow$ & $M$ & $S\downarrow$ & $T_f\uparrow$ & $T_r\uparrow$ & $M$ \\
\midrule
\multirow{6}{*}{\shortstack[l]{\emph{Source-blind}\\\emph{correction}}} & Identity & 0.989 & 0.728 & 0.728 & 0.000 & 0.916 & 0.688 & 0.688 & 0.000 \\
 & LEACE & 0.980 & 0.728 & 0.727 & 0.234 & 0.918 & 0.689 & 0.689 & 0.102 \\
 & \textbf{MGPA-CF} & 0.982 & 0.727 & 0.728 & 0.237 & 0.895 & 0.689 & 0.690 & 0.056 \\
 & \textbf{MGPA-Iter} & 0.946 & 0.735 & 0.746 & 0.347 & 0.872 & 0.688 & 0.689 & 0.149 \\
 & \quad Linear-only ablation & 0.987 & 0.725 & 0.729 & 0.345 & 0.872 & 0.689 & 0.689 & 0.069 \\
 & IGBP & 0.955 & 0.722 & 0.710 & 0.497 & 0.872 & 0.689 & 0.688 & 0.096 \\
\midrule
\multirow{4}{*}{\shortstack[l]{\emph{Source-aware}\\\emph{alignment}}} & CORAL $\to$ ref. 0 & 0.948 & 0.738 & 0.735 & 0.448 & 1.000 & 0.691 & 0.688 & 0.149 \\
 & CORAL $\to$ ref. 1 & 0.958 & 0.726 & 0.740 & 0.292 & 1.000 & 0.685 & 0.688 & 0.155 \\
 & FEATMAP $\to$ ref. 0 & 0.880 & 0.777 & 0.749 & 0.800 & N/A & N/A & N/A & N/A \\
 & FEATMAP $\to$ ref. 1 & 0.890 & 0.693 & 0.711 & 0.576 & N/A & N/A & N/A & N/A \\
\bottomrule
\end{tabular*}\endgroup

\end{table}

\textbf{Recorded SSVEP.}
\label{sec:recorded-result}%
Both constructions shift the pooled token mean, while readers and the task head see all 15 tokens (Appendices~\ref{app:full-token-readers} and~\ref{app:recorded-full-input}). Both reduce full-input source accessibility relative to LEACE while leaving task scores close to identity (Table~\ref{tab:real-main}B): the paired source differences are $-.0231$ $[-.0459,-.0058]$ for Closed-form and $-.0463$ $[-.0671,-.0225]$ for Iterative MGPA. Iterative MGPA, its linear-only ablation and IGBP share $S=.872$, set by readers of unchanged token residuals, and IGBP uses less movement ($.096$ versus $.149$) with close task AUROCs.

\subsection{Cross-task reuse under device--label association shifts}
\label{sec:recovery}
This study tests whether one map selected on P300 improves existing N170 and MMN heads. All three tasks share simultaneous
Neuroscan/Flex recordings, one EEGPT/PCA representation and the same eight
evaluation participants; encoder and heads stay frozen. For each task, a
shortcut-sensitive head is trained under device--label association and
evaluated as that association weakens or reverses, by selecting the
observed device from real event pairs without altering the EEG; a
separately trained control head uses independent assignment. P300
development data select correction strength by worst-association AUROC
under task-utility constraints, with the same rule for competitors
(Appendix~\ref{app:recovery-protocol}).

\begin{table}[t]
\centering\small
\caption{\textbf{One P300-selected correction, three frozen task heads.}
Eight common participants. Worst: minimum participant-mean AUROC of the
shortcut-trained head over device--label association strength; Control:
separately trained head with device assignment independent of the label.
Paired intervals and per-fit results: Appendix~\ref{app:transfer-results}.}
\label{tab:reuse-main}
\begingroup\small\setlength{\tabcolsep}{4pt}
\begin{tabular*}{\textwidth}{@{\extracolsep{\fill}}llrrrrrr@{}}
\toprule
 & & \multicolumn{3}{c}{Worst-association AUROC $\uparrow$} & \multicolumn{3}{c}{Control-head AUROC $\uparrow$} \\
\cmidrule(lr){3-5}\cmidrule(lr){6-8}
Interface & Method & P300 & N170 & MMN & P300 & N170 & MMN \\
\midrule
\multirow{4}{*}{\shortstack[l]{\emph{Source-blind}\\\emph{correction}}} & Identity & 0.355 & 0.122 & 0.446 & 0.627 & 0.622 & 0.640 \\
 & LEACE & 0.601 & 0.384 & 0.614 & 0.622 & 0.621 & 0.639 \\
 & \textbf{MGPA-CF} & 0.599 & 0.420 & 0.642 & 0.627 & 0.621 & 0.642 \\
 & \textbf{MGPA-Iter} & 0.599 & 0.441 & 0.632 & 0.627 & 0.621 & 0.641 \\
\midrule
\multirow{2}{*}{\shortstack[l]{\emph{Source-aware}\\\emph{alignment}}} & CORAL & 0.550 & 0.277 & 0.624 & 0.627 & 0.622 & 0.641 \\
 & FEATMAP & 0.573 & 0.471 & 0.618 & 0.627 & 0.611 & 0.625 \\
\bottomrule
\end{tabular*}\endgroup

\end{table}

Both MGPA constructions improve worst-association AUROC over LEACE on the
two transfer tasks (Table~\ref{tab:reuse-main}; Figure~\ref{fig:reuse} in
Appendix~\ref{app:transfer-results} adds marginal intervals).
Iterative MGPA gains $.0574$ $[.0338,.0831]$ on N170 and $.0188$
$[.0040,.0241]$ on MMN; Closed-form MGPA gains $.0364$ $[.0049,.0667]$ and
$.0282$ $[.0021,.0343]$, respectively. Separate control-head AUROC stays
close to identity across all three tasks. FEATMAP has the highest N170 worst-association score, with lower N170 and MMN control-head scores and a source-aware deployment interface.
The MGPA gain costs performance under the training (aligned) association relative to the uncorrected head: on N170, Iterative MGPA changes aligned AUROC from $.914$ to $.744$ and worst AUROC from $.122$ to $.441$, still below chance but two thirds of the way to the control head ($.621$).
\section{Discussion and Conclusion}
\label{sec:discussion}
Source predictability is a symptom, not a license to edit: only the part the measurement caused is an artifact, and erasure cannot tell it from the rest. MGPA therefore splits correction into two decisions: calibration constrains the directions that can change, and conditional anchoring accounts for the source evidence in the coordinates that remain fixed. Our analysis shows that, for a fixed affine score, this anchor retains the same information as any other target set by the preserved coordinates at the least expected squared movement, and characterizes selective erasure in a Gaussian model. In practice, a frozen encoder and its existing heads can stay in use across headset, electrode or reference changes, with short paired calibration, no retraining and no source labels at deployment.

The experiments support both choices. In controlled SSVEP, where source is tied to the task, MGPA removes source without the task loss that LEACE incurs; ablations attribute the attenuation to the gate's directions and the lower movement to the conditional target. On recorded EEG, Iterative MGPA lowers source accessibility with task performance preserved or improved, and nonlinear critics add to this on temporal N170 features. A P300-selected map, reused unchanged, makes frozen N170 and MMN heads more robust to device--label association shifts than LEACE does, trading part of their aligned performance for that robustness. Readers and existing heads measure different things (Section~\ref{sec:protocol}), so corrected representations should be evaluated through both.

The present evidence has limits. Calibration identifies responses to the supplied measurement contrasts; it does not isolate task-free directions or guarantee coverage of other acquisition changes, which matters most for the virtual contrasts used for recorded SSVEP. The theoretical guarantees apply to the stated affine and Gaussian settings, not to refitted nonlinear trajectories. Cross-task reuse is tested on one device pair with eight evaluation participants, so broader calibration and larger cohorts are needed. Within these limits, the construction is not EEG-specific: \emph{wherever credible views of one observation under different measurement conditions exist, measurement evidence can decide where a correction may act, and preserved information what it should aim for.}

\FloatBarrier
\label{sec:main-end}
\section*{Reproducibility Statement}
The supplement provides proofs, calibration and update specifications,
participant roles, selection rules and statistical evaluation details.
Code, configurations and participant-level results are indexed by the
repository README at \url{https://github.com/sneddy/mgpa-paper}, which distinguishes
saved-result reconstruction from fresh experiments.

\section*{Ethics Statement}
This study uses existing research EEG datasets and introduces no new data
collection or clinical intervention. Provenance attenuation does not confer
anonymity or establish clinical validity. Applications involving patients
require task-specific validation and appropriate data governance.

\section*{AI Usage Statement}
Generative artificial intelligence (AI) tools assisted code cleaning and
refactoring, preliminary literature search, and manuscript text polishing.
All AI-assisted code was reviewed and tested by the authors. The authors
retain full responsibility for the final content.

\section*{Acknowledgments}
We would like to acknowledge support from the MBZUAI Startup Fund.

\bibliography{references}

@inproceedings{wang2024eegpt,
  author = {Guangyu Wang and Wenchao Liu and Yuhong He and Cong Xu and Lin Ma and Haifeng Li},
  title = {{EEGPT}: Pretrained Transformer for Universal and Reliable Representation of {EEG} Signals},
  booktitle = {Advances in Neural Information Processing Systems},
  volume = {37},
  pages = {39249--39280},
  year = {2024},
  doi = {10.52202/079017-1239},
}

@book{nunez2006,
  author = {Paul L. Nunez and Ramesh Srinivasan},
  title = {Electric Fields of the Brain: The Neurophysics of {EEG}},
  edition = {Second},
  publisher = {Oxford University Press},
  year = {2006},
  isbn = {9780195050387},
  doi = {10.1093/acprof:oso/9780195050387.001.0001},
}

@article{dong2019,
  author = {Li Dong and Xiaobo Liu and Lingling Zhao and Yongxiu Lai and Diankun Gong and Tiejun Liu and Dezhong Yao},
  title = {A Comparative Study of Different {EEG} Reference Choices for Event-Related Potentials Extracted by Independent Component Analysis},
  journal = {Frontiers in Neuroscience},
  volume = {13},
  pages = {1068},
  year = {2019},
  doi = {10.3389/fnins.2019.01068},
}

@article{geirhos2020,
  author = {Robert Geirhos and J{\"o}rn-Henrik Jacobsen and Claudio Michaelis and Richard Zemel and Wieland Brendel and Matthias Bethge and Felix A. Wichmann},
  title = {Shortcut Learning in Deep Neural Networks},
  journal = {Nature Machine Intelligence},
  volume = {2},
  number = {11},
  pages = {665--673},
  year = {2020},
  doi = {10.1038/s42256-020-00257-z},
}

@article{zech2018,
  author = {John R. Zech and Marcus A. Badgeley and Manway Liu and Anthony B. Costa and Joseph J. Titano and Eric Karl Oermann},
  title = {Variable Generalization Performance of a Deep Learning Model to Detect Pneumonia in Chest Radiographs: A Cross-Sectional Study},
  journal = {{PLOS} Medicine},
  volume = {15},
  number = {11},
  pages = {e1002683},
  year = {2018},
  doi = {10.1371/journal.pmed.1002683},
}

@article{yamashita2019,
  author = {Ayumu Yamashita and Noriaki Yahata and Takashi Itahashi and Giuseppe Lisi and Takashi Yamada and Naho Ichikawa and Masahiro Takamura and Yujiro Yoshihara and Akira Kunimatsu and Naohiro Okada and Hirotaka Yamagata and Koji Matsuo and Ryuichiro Hashimoto and Go Okada and Yuki Sakai and Jun Morimoto and Jin Narumoto and Yasuhiro Shimada and Kiyoto Kasai and Nobumasa Kato and Hidehiko Takahashi and Yasumasa Okamoto and Saori C. Tanaka and Mitsuo Kawato and Okito Yamashita and Hiroshi Imamizu},
  title = {Harmonization of Resting-State Functional {MRI} Data across Multiple Imaging Sites via the Separation of Site Differences into Sampling Bias and Measurement Bias},
  journal = {{PLOS} Biology},
  volume = {17},
  number = {4},
  pages = {e3000042},
  year = {2019},
  doi = {10.1371/journal.pbio.3000042},
}

@inproceedings{ravfogel2020,
  author = {Shauli Ravfogel and Yanai Elazar and Hila Gonen and Michael Twiton and Yoav Goldberg},
  title = {Null It Out: Guarding Protected Attributes by Iterative Nullspace Projection},
  booktitle = {Proceedings of the 58th Annual Meeting of the Association for Computational Linguistics},
  pages = {7237--7256},
  year = {2020},
  doi = {10.18653/v1/2020.acl-main.647},
}

@inproceedings{belrose2023,
  author = {Nora Belrose and David Schneider-Joseph and Shauli Ravfogel and Ryan Cotterell and Edward Raff and Stella Biderman},
  title = {{LEACE}: Perfect Linear Concept Erasure in Closed Form},
  booktitle = {Advances in Neural Information Processing Systems},
  volume = {36},
  pages = {66044--66063},
  year = {2023},
  doi = {10.52202/075280-2884},
}

@inproceedings{iskander2023,
  author = {Shadi Iskander and Kira Radinsky and Yonatan Belinkov},
  title = {Shielded Representations: Protecting Sensitive Attributes through Iterative Gradient-Based Projection},
  booktitle = {Findings of the Association for Computational Linguistics: ACL 2023},
  pages = {5961--5977},
  year = {2023},
  doi = {10.18653/v1/2023.findings-acl.369},
}

@misc{avitan2026mance,
  author = {Matan Avitan and Yoav Goldberg and Yanai Elazar},
  title = {{MANCE}: Manifold Aware Concept Erasure},
  year = {2026},
  eprint = {2607.03973},
  archiveprefix = {arXiv},
  primaryclass = {cs.LG},
  doi = {10.48550/arXiv.2607.03973},
  url = {https://arxiv.org/abs/2607.03973}
}

@misc{bai2025ncm,
  author = {Ruqi Bai and Yao Ji and Zeyu Zhou and David I. Inouye},
  title = {From Invariant Representations to Invariant Data: Provable Robustness to Spurious Correlations via Noisy Counterfactual Matching},
  year = {2025},
  eprint = {2505.24843},
  archiveprefix = {arXiv},
  primaryclass = {cs.LG},
  doi = {10.48550/arXiv.2505.24843},
  url = {https://arxiv.org/abs/2505.24843}
}

@inproceedings{pogodin2023circe,
  author = {Roman Pogodin and Namrata Deka and Yazhe Li and Danica J. Sutherland and Victor Veitch and Arthur Gretton},
  title = {Efficient Conditionally Invariant Representation Learning},
  booktitle = {International Conference on Learning Representations},
  year = {2023},
  url = {https://openreview.net/forum?id=dJruFeSRym1}
}

@inproceedings{hewitt2021conditional,
  author = {John Hewitt and Kawin Ethayarajh and Percy Liang and Christopher Manning},
  title = {Conditional Probing: Measuring Usable Information Beyond a Baseline},
  booktitle = {Proceedings of the 2021 Conference on Empirical Methods in Natural Language Processing},
  pages = {1626--1639},
  year = {2021},
  doi = {10.18653/v1/2021.emnlp-main.122},
}

@inproceedings{ngweta2023,
  author = {Lilian Ngweta and Subha Maity and Alex Gittens and Yuekai Sun and Mikhail Yurochkin},
  title = {Simple Disentanglement of Style and Content in Visual Representations},
  booktitle = {Proceedings of the 40th International Conference on Machine Learning},
  series = {Proceedings of Machine Learning Research},
  volume = {202},
  pages = {26063--26086},
  year = {2023},
  publisher = {PMLR},
  url = {https://proceedings.mlr.press/v202/ngweta23a.html}
}

@inproceedings{gnassounou2023,
  author = {Th{\'e}o Gnassounou and R{\'e}mi Flamary and Alexandre Gramfort},
  title = {Convolution {Monge} Mapping Normalization for Learning on Sleep Data},
  booktitle = {Advances in Neural Information Processing Systems},
  volume = {36},
  pages = {10457--10476},
  year = {2023},
  url = {https://proceedings.neurips.cc/paper_files/paper/2023/hash/21718991f6acf19a42376b5c7a8668c5-Abstract-Conference.html}
}

@article{zhu2021ssvep,
  author = {Fangkun Zhu and Lu Jiang and Guoya Dong and Xiaorong Gao and Yijun Wang},
  title = {An Open Dataset for Wearable {SSVEP}-Based Brain--Computer Interfaces},
  journal = {Sensors},
  volume = {21},
  number = {4},
  pages = {1256},
  year = {2021},
  doi = {10.3390/s21041256},
}

@article{williams2020flex,
  author = {Williams, Nikolas S. and McArthur, Genevieve M. and de Wit, Bianca and Ibrahim, George and Badcock, Nicholas A.},
  title = {A Validation of {Emotiv} {EPOC} {Flex} Saline for {EEG} and {ERP} Research},
  journal = {PeerJ},
  volume = {8},
  pages = {e9713},
  year = {2020},
  doi = {10.7717/peerj.9713},
}

@article{more1983,
  author = {Jorge J. Mor{\'e} and D. C. Sorensen},
  title = {Computing a Trust Region Step},
  journal = {SIAM Journal on Scientific and Statistical Computing},
  volume = {4},
  number = {3},
  pages = {553--572},
  year = {1983},
  doi = {10.1137/0904038},
}

@article{donle2026featmap,
  author = {Leonhard Donle and Michael Phillips and Farieda Gaber and Siddhi Ramesh and Matteo Sacco and Sampsa Hautaniemi and Anni Virtanen and Keno Bressem and Lisa Adams and Kelsey Goon and Elena Nevins and Ryan A. Robinett and Sara Kochanny and Sasha Hassan and James Dolezal and Alexander T. Pearson and Ernst Lengyel},
  title = {{FEATMAP}: Targeted Correction of Acquisition Signatures Harmonizes Medical Foundation Model Embeddings and Enables Robust Task Generalization},
  journal = {bioRxiv},
  year = {2026},
  doi = {10.64898/2026.07.02.736184},
}

@article{eikenberry2026,
  author = {Keenan Eikenberry and Lizuo Liu and Yoonsang Lee},
  title = {Post-Training Augmentation Invariance},
  journal = {Transactions on Machine Learning Research},
  year = {2026},
  url = {https://openreview.net/forum?id=Z4uUwU6zRe}
}

@inproceedings{sun2016coral,
  author = {Baochen Sun and Jiashi Feng and Kate Saenko},
  title = {Return of Frustratingly Easy Domain Adaptation},
  booktitle = {Proceedings of the AAAI Conference on Artificial Intelligence},
  volume = {30},
  pages = {2058--2065},
  year = {2016},
  doi = {10.1609/aaai.v30i1.10306},
}

@article{he2020ea,
  author = {He He and Dongrui Wu},
  title = {Transfer Learning for Brain--Computer Interfaces: A {Euclidean} Space Data Alignment Approach},
  journal = {IEEE Transactions on Biomedical Engineering},
  volume = {67},
  number = {2},
  pages = {399--410},
  year = {2020},
  doi = {10.1109/TBME.2019.2913914},
}

@article{rodrigues2019rpa,
  author = {Pedro Luiz Coelho Rodrigues and Christian Jutten and Marco Congedo},
  title = {Riemannian {Procrustes} Analysis: Transfer Learning for Brain--Computer Interfaces},
  journal = {IEEE Transactions on Biomedical Engineering},
  volume = {66},
  number = {8},
  pages = {2390--2401},
  year = {2019},
  doi = {10.1109/TBME.2018.2889705},
}

@article{wu2016awar,
  author = {Dongrui Wu and Vernon J. Lawhern and W. David Hairston and Brent J. Lance},
  title = {Switching {EEG} Headsets Made Easy: Reducing Offline Calibration Effort Using Active Weighted Adaptation Regularization},
  journal = {IEEE Transactions on Neural Systems and Rehabilitation Engineering},
  volume = {24},
  number = {11},
  pages = {1125--1137},
  year = {2016},
  doi = {10.1109/TNSRE.2016.2544108},
}

@article{liu2026sdda,
  author = {Dingkun Liu and Siyang Li and Ziwei Wang and Wei Li and Dongrui Wu},
  title = {{SDDA}: Spatial Distillation Based Distribution Alignment for Cross-Headset {EEG} Classification},
  journal = {IEEE Transactions on Biomedical Engineering},
  volume = {73},
  number = {7},
  pages = {2400--2411},
  year = {2026},
  doi = {10.1109/TBME.2025.3631604},
}

@inproceedings{holstege2025splince,
  title     = {Preserving Task-Relevant Information Under Linear Concept Removal},
  author    = {Holstege, Floris and Ravfogel, Shauli and Wouters, Bram},
  booktitle = {Advances in Neural Information Processing Systems},
  volume    = {38},
  year      = {2025},
  doi       = {10.52202/085713-0907},
}
\bibliographystyle{iclr2027_conference}
\clearpage
\appendix
\section{Proofs and Solver Details}
\label{app:proofs}
All statements in this section condition on the fitted gate and treat its
orthonormal basis $B$ as fixed. Fixed-score statements also condition on the
critic and correction metric. Norms without a subscript are Euclidean;
$\|v\|_\Omega^2=v^\top\Omega v$. Conditioning on fitted quantities means that they are treated as fixed when averaging over evaluation observations. We use $r$ for the editable dimension and $d$ for the full representation dimension; each identity matrix $I$ has the dimension of the space on which it acts.

\subsection{Fixed-complement preservation and the source floor}
\label{app:fixed-complement}
Since $(I-P)B=0$, every admissible correction satisfies
\[
 (I-P)R(h)=(I-P)h=q(h).
\]
Thus $Q$ is recoverable from the corrected representation, and the chain
rule gives \eqref{eq:floor}. Preservation also holds under composition and
fractional steps. To contrast selective edits with an output determined entirely by preserved coordinates, let $\mu(q)\in\R^r$ be a measurable choice of editable coordinates. This vector-valued completion is local to this subsection; it is distinct from the scalar mean-score anchor $\mu(q)$ in Proposition~\ref{prop:parallel-levels}. For $R_\mu(h)=q(h)+B\mu(q(h))$, the output is determined
by $Q$, so the conditional-information term is zero.

\textbf{Bayes-score decomposition.}\label{app:bayes-split}
Let $s(q,z)$ be the Bayes source score, the posterior log-odds of source $1$ versus source $0$ given both coordinates, and let $P_{Z\mid Q=q,U=u}$ denote the conditional law of $Z$ for source $u$ at fixed $q$; this subscripted $P$ is a distribution, not the gate matrix $P=BB^\top$. For $0<p(q)<1$, write $\sg(q)=\logit p(q)$ with $\logit p=\log\{p/(1-p)\}$. When the two conditional laws are mutually absolutely continuous, that is, they assign zero probability to the same sets, the finite Bayes source score under the original representation distribution satisfies
\begin{equation}
 s(q,z)=\sg(q)+\log\frac{dP_{Z\mid Q=q,U=1}}{dP_{Z\mid Q=q,U=0}}(z).
 \label{eq:odds}
\end{equation}
The Radon--Nikodym derivative in \eqref{eq:odds} is the likelihood ratio between these laws; when densities exist, it is their density ratio evaluated at $z$. Its logarithm is the source evidence supplied by $Z$ beyond $Q$. Setting the original score to zero therefore requires the editable contribution to cancel $\sg(q)$, although $Q$, and with it that evidence, remains recoverable after any admissible correction. The fixed affine score of Proposition~\ref{prop:parallel-levels} need not have this form: the decomposition motivates the target choice, and the proposition does not rely on it.

\subsection{Fixed-score geometry and retained information}
\label{app:affine-levels}
\label{app:euclidean-anchor}
In Proposition~\ref{prop:parallel-levels}, the functions $b(q),w(q)$ and
$\Omega(q)\succ0$ are measurable in $q$, with $w(q)\ne0$ almost surely.
The score has finite conditional second moments when costs are evaluated. Let $\Delta\in\R^r$ denote the displacement added to $z$; dependence of $b,w,\Omega$ and $v$ on the fixed $q$ is suppressed below.
For fixed $(q,z)$, matching anchor $a(q)$ requires
$w^\top\Delta=a(q)-s(q,z)$. Metric Cauchy--Schwarz gives
\[
 (a-s)^2\le(w^\top\Omega^{-1}w)\,\Delta^\top\Omega\Delta.
\]
The unique minimizer is $\Delta=v(a-s)$, where
$v=\Omega^{-1}w/d_\Omega$. Its squared norm is $(a-s)^2/d_\Omega$.
The output editable block is
\[
 Z'=N+v\{a(Q)-b(Q)\},\qquad N=(I-vw^\top)Z.
\]
The output is a measurable function of $(Q,N)$; conversely,
$N=Z'-v\{a(Q)-b(Q)\}$ is measurable in the output because $Q$ is
retained. Thus $\sigma(T_a)=\sigma(Q,N)$. Here $\sigma(T_a)$ abbreviates the sigma-algebra of the random output $T_a(Q,Z)$, rather than of an uninstantiated map. Every anchor retains the same
information about $U$ and about any task variable $Y$, including the same
unrestricted Bayes prediction risk. Conditional independence from $U$
holds exactly when $N\perp U\mid Q$. The conditional bias--variance
decomposition of $\E[(s-a)^2\mid Q=q]$ proves
\eqref{eq:affine-cost} and the unique mean-anchor optimum.

For $\Omega=I$, the edit is $w(a-s)/\|w\|^2$ and the denominator is
$\|w\|^2$. Orthonormality of $B$ makes this the Euclidean movement in
the standardized representation as well. The result compares exact
score-level corrections with the same fixed score and metric. Nonlinear
sample-dependent Jacobians, active radius constraints and independently
refitted compositions are not members of that comparison.

\subsection{Proof of Theorem~\ref{thm:compensation}}
\label{app:gaussian-proof}
Fix $q$ and abbreviate $p=p(q)$, $m=m(q)$ and $\kappa=\kappa(q)$. Throughout this proof, conditional moments are taken at this fixed $q$. In the ratio below, $p(z\mid q,u)$ denotes the conditional density of $Z$, whereas the scalar $p$ is the conditional probability of source $1$.
Expanding the two Gaussian quadratic log densities gives
\[
 \log\frac{p(z\mid q,1)}{p(z\mid q,0)}
 =\delta^\top \Sigma^{-1}(z-m)=\ell(q,z).
\]
Let $L=I-\delta\delta^\top \Sigma^{-1}/D$. This $r\times r$ matrix is the linear part of the correction and annihilates the source-mean shift. The corrected editable random vector $Z_\kappa$ is
\[
 Z_\kappa=LZ+\frac{\delta}{D}
                  \{\delta^\top \Sigma^{-1}m+\kappa\}.
\]
Since $L\delta=0$, its conditional mean is the same for both sources.
Its conditional covariance is $L\Sigma L^\top$ for both. The resulting possibly
singular Gaussian laws are identical, proving conditional independence.
The map preserves $Q$ by construction.

Proposition~\ref{prop:parallel-levels}, applied with $w=\Sigma^{-1}\delta$ and
$\Omega=\Sigma^{-1}$, gives the unique minimum-norm edit
$\Delta=\delta(\kappa-\ell)/D$, with squared cost $(\kappa-\ell)^2/D$.
Conditional on $Q=q,U=u$, the residual score $\ell$ has mean
$(2u-1)D/2$ and variance $D$. Consequently,
\[
 \E[\ell\mid q]=(2p-1)D/2,\qquad
 \Var(\ell\mid q)=D+D^2p(1-p).
\]
Thus
\[
 \mathcal C_\kappa(q)=\frac{\E[(\ell-\kappa)^2\mid q]}{D}
 =1+Dp(1-p)+\frac{\{\kappa-(2p-1)D/2\}^2}{D},
\]
which is \eqref{eq:gauge-cost}. Its minimum occurs at
$\kappa_*=(2p-1)D/2$. Subtracting this minimum from the cost at
$\kappa=-\sg$ yields \eqref{eq:compensation}. Since $\sg=\logit p$ and
$2p-1$ have the same sign, the excess is strictly positive for $p\ne1/2$.
At fixed such $p$, its numerator tends to $\sg^2>0$ as $D\downarrow0$,
so the excess diverges.

\subsection{Selective mean anchoring and full completion}
\label{app:completion}
The unique minimizer of \eqref{eq:gauge-cost} is
$\kappa_*=(2p-1)D/2$. To show its optimality among deterministic conditional
erasers affine in $Z$, whiten at fixed $q$. In the following change of coordinates, $X\in\R^r$ is the centered, whitened editable vector, and $\Sigma^{-1/2}$ is the symmetric inverse square root of the covariance. Within this subsection only, $v\in\R^r$ denotes the whitened mean shift, rather than the normalized projection direction in \eqref{eq:affine-map}:
\[
 X=\Sigma^{-1/2}(Z-m),\quad v=\Sigma^{-1/2}\delta,\quad
 X\mid U=u\sim\mathcal N((2u-1)v/2,I),\quad\|v\|^2=D.
\]
We compare affine maps with linear part $F\in\R^{r\times r}$ and offset $c\in\R^r$. An affine map $X'=FX+c$ erases source exactly if and only if $Fv=0$:
the two output means must coincide, and their covariance already agrees.
Here $\E X$ and $\Cov(X)$ are the pooled mean and covariance at fixed $q$, averaging over both sources. We write $\tr$ for matrix trace and $\|\cdot\|_F$ for the Frobenius norm in the cost comparison below.
For fixed $F$, the least-distortion intercept is $c=(I-F)\E X$. Since
$\Cov(X)=I+p(1-p)vv^\top$, its cost under the erasure constraint is
\[
 \tr\{(F-I)\Cov(X)(F-I)^\top\}
 =\|F-I\|_F^2+p(1-p)D,\qquad Fv=0.
\]
The unique matrix nearest $I$ in Frobenius norm subject to $Fv=0$ is
$F_*=I-vv^\top/D$. Unwhitening gives precisely $R_{\kappa_*}$, with
cost $1+Dp(1-p)$. This is the conditional binary affine-erasure geometry
of LEACE \citep{belrose2023}.

For quotient-only completion, the editable output depends only on the preserved $Q$. The best fixed-$q$ choice is
$Z'=\E[Z\mid q]$. Its cost is the $\Sigma^{-1}$ trace of
$\Cov(Z\mid q)=\Sigma+p(1-p)\delta\delta^\top$, hence
$r+Dp(1-p)$, exceeding the selective correction cost by $r-1$.

\subsection{Partial closed-form steps}
\label{app:partial-step}
On the editable block, a full closed-form step has linear part
$M=I-\widehat\delta\widehat\delta^\top C^{-1}/(\widehat\delta^\top C^{-1}\widehat\delta)$,
where $I-M$ is a rank-one idempotent. A step of strength $\alpha$ has linear part
$M_\alpha=I-\alpha(I-M)$ with $\det M_\alpha=1-\alpha$. Thus for $\alpha\ne1$ the
map is invertible given $Q$ and preserves the exact conditional source information
and covariance-adjusted separation, even though a fixed prediction head can respond
differently.

\subsection{Estimating the same conditional-mean anchor}
\label{app:anchor-estimation}
Fix the gate, affine score, metric and evaluation distribution in
Proposition~\ref{prop:parallel-levels}. Let an anchor-fitting sample
$\mathcal C$, independent of the evaluation observation, produce a
$Q$-dependent anchor $\widehat a_{\mathcal C}(q)$ with finite mean-squared error.
Here $\mathcal C$ without a subscript denotes the fitting sample, whereas $\mathcal C_a(q)$ remains the conditional movement cost for target $a$. The operator $\E_{\mathcal C}$ averages over possible fitting samples; the cost itself averages over evaluation observations at fixed $q$.
Writing $\mu(q)=\E[s\mid Q=q]$, Proposition~\ref{prop:parallel-levels}
and expectation over $\mathcal C$ give
\begin{equation}
 \E_{\mathcal C}\mathcal C_{\widehat a_{\mathcal C}}(q)-\mathcal C_\mu(q)
 =\frac{\E_{\mathcal C}\{\widehat a_{\mathcal C}(q)-\mu(q)\}^2}{d_\Omega(q)}.
 \label{eq:anchor-risk}
\end{equation}
Thus the estimated anchor has lower expected squared movement than a fixed
reference $a_0(q)$ exactly when its estimation mean-squared error is below
$\{a_0(q)-\mu(q)\}^2$. This is an out-of-sample comparison conditional
on the fixed score and gate.

\subsection{The smooth trust-region step}
\label{app:solver}
For $J$ critics, let $A\in\R^{J\times r}$ and $e\in\R^J$ be the editable Jacobian and score-residual vector from Section~\ref{sec:method}. Row $A_j$ is the gated gradient of critic $j$, with $j=1,\ldots,J$. A diagonal matrix $W\in\R^{J\times J}$ balances these gradient scales using the positive stabilizer $\lambda$:
\begin{equation}
 W_{jj}=(\|A_j\|^2+\lambda^2)^{-1/2},\qquad
 \bar A=WA,\quad\bar e=We,\qquad\lambda>0.
 \label{eq:normalization}
\end{equation}
The bars denote the weighted Jacobian and residual. Recall that $\tau$ penalizes squared movement and $\rho$ is the maximum editable displacement per stage.
For $\lambda,\tau,\rho>0$, the objective in \eqref{eq:tr-problem} is
strictly convex on a compact convex ball. Its unique minimizer has the
standard trust-region form \citep{more1983}. In the expression below, $\nu\ge0$ is the Lagrange multiplier for the radius constraint, and the inverse acts on a $J\times J$ matrix:
\begin{equation}
 \Delta z_*=-\bar A^\top\{\bar A\bar A^\top+(\tau+\nu)I\}^{-1}\bar e,
 \quad\nu\ge0,\quad\nu(\|\Delta z_*\|-\rho)=0.
 \label{eq:tr-solution}
\end{equation}
The multiplier can be positive only when the step reaches the radius. The equivalent stationarity condition acts in the $r$-dimensional editable space, so its identity matrix is $r\times r$:
\[
 (\bar A^\top\bar A+(\tau+\nu)I)\Delta z_*=-\bar A^\top\bar e
\]
This condition yields the stated $J\times J$ solution. Set $\nu=0$ when the unconstrained
step is feasible; otherwise its norm decreases continuously with $\nu$,
and a scalar solve enforces $\|\Delta z_*\|=\rho$. Since $B$ is orthonormal,
the radius directly bounds the embedding edit:
$\|h^+-h\|=\|\Delta z_*\|\le\rho$.

\section{Experimental Protocols}
\label{app:protocols}
\label{app:protocol}
Each split assigns participants to disjoint roles: FIT estimates coordinates,
gates and adapters; VAL determines prescribed movement endpoints; HEAD
trains independent source readers and task heads; EVAL provides participant
scores. These names denote fitting, validation, head-training and evaluation
sets, respectively. In reuse, the P300 development set DEV replaces VAL;
CAL denotes paired calibration observations drawn from FIT.
Table~\ref{tab:protocol-map} distinguishes native outputs, matched movement
and P300 DEV selection. Downstream labels construct associations
and train heads, but enter adapter selection only in reuse and never
adapter fitting. The repository README (\url{https://github.com/sneddy/mgpa-paper})
indexes participant lists, executable configurations and reproduction
instructions for each study.

\begin{table}[tp]
\centering\small
\caption{\textbf{Protocol map.} Participant counts are FIT/VAL/HEAD/EVAL
(DEV replaces VAL in reuse); CAL denotes the calibration pairs. CAR is
common-average referencing, Oz is the reference electrode, and PCA256
retains 256 principal components. Gates are fixed during correction.}
\label{tab:protocol-map}
\begin{tabular}{@{}>{\raggedright\arraybackslash}p{.18\linewidth}>{\raggedright\arraybackslash}p{.43\linewidth}>{\raggedright\arraybackslash}p{.33\linewidth}@{}}
\toprule
Study & Coordinates, calibration and gate & Participants and operating point \\
\midrule
Controlled SSVEP & Pooled EEGPT512; full same-trial CAR/Oz contrasts;
90\% directional energy, ranks $3,3,3,2$ & Four role rotations;
81 distinct EVAL participants. Native correction/target; $.05$ VAL budget for permission. \\
\addlinespace
Temporal N170 & 240 temporal features; simultaneous Neuroscan/Flex pairs;
90\% energy, rank 56 & $4/2/4/10$; native maps and matched-budget IGBP comparison. \\
\addlinespace
Recorded SSVEP & Mean512 correction, $15\times512$ readouts;
six virtual FIT-local contrasts; 99\% energy, rank 328 & $41/20/20/21$;
native maps. \\
\addlinespace
P300-selected reuse & EEGPT/PCA256, P300 FIT reference coordinates;
400 simultaneous CAL pairs, rank 8 & $4/2/4/8$;
P300 DEV-selected map, unchanged on N170/MMN. \\
\bottomrule
\end{tabular}
\end{table}

\subsection{Measurement views and participant roles}
\textbf{Controlled SSVEP.}
The wet/dry collection contains 102 participants, 12 frequencies and ten
blocks per frequency and electrode type \citep{zhu2021ssvep}. Wet trials
from eight occipital channels are cropped to samples 160--659 at 250 Hz,
standardized within channel and encoded by frozen EEGPT
\citep{wang2024eegpt}. We encode common-average-reference (CAR) and
Oz-electrode-reference views separately. For underlying trial $i$,
averaging 15 tokens yields $h_i^{\rm CAR},h_i^{\rm Oz}\in\R^{512}$,
where the superscripts identify the reference condition. The binary index
$u$ selects the evaluated view, and $\gamma$ scales its contrast:
\[
 h_i(u)=\tfrac12(h_i^{\rm CAR}+h_i^{\rm Oz})
 +(2u-1)\tfrac{\gamma}{2}(h_i^{\rm Oz}-h_i^{\rm CAR}),\qquad\gamma=.005,
\]
Here $u=0$ selects the scaled CAR view and $u=1$ the scaled Oz view,
whereas calibration uses the full CAR--Oz contrasts. The task is frequency
$\geq12$ Hz. Within each FIT/VAL participant--frequency cell, nine of ten
trials receive source 1 in the high-frequency class and one of ten in the
low-frequency class. Adapters fit these observed endpoints; HEAD/EVAL
readers use both endpoints with equal source mass. This association emulates the second route in
Figure~\ref{fig:measurement-paths}: source labels alone confound acquisition
with task, so a label-fitted eraser such as LEACE also removes task
directions. MGPA takes only its gate and Closed-form direction from
calibration pairs; its anchors, covariance metric and critics are fitted on
the same associated rows as LEACE.

Five folds are stratified by electrode-session order and gender. Here $r$
indexes the role rotation, independently of the gate-rank notation. In
rotation $r$, folds $r,r+1,r+2$ supply EVAL, HEAD, VAL cyclically, and the
remaining two supply FIT. The four FIT/VAL/HEAD/EVAL counts are $41/20/20/21$,
$42/20/20/20$, $41/21/20/20$ and $40/21/21/20$. Each participant supplies
120 underlying trials. The 81 distinct EVAL participants appear once per
fit seed; source assignment is fixed across seeds. Closed-form comparisons
use one deterministic map per split and the seed-17 reader realization;
gate/target comparisons cross these rotations with all shared fit seeds.

\textbf{Temporal N170.}
The public Neuroscan/Flex dataset \citep{williams2020flex} captures every
watch/face event on both devices at once; Neuroscan is source 0, Flex is
source 1, and the two views are matched by event ordinal and timestamp. Both
views keep the 16 channels the devices share, are band-passed at 0.1--30 Hz
with a zero-phase FIR filter, re-referenced to the channel average,
downsampled to 128 Hz and cut into 1 s windows from $-.2$ s with the
prestimulus mean removed. We drop a pair when either view exceeds
$100\,\mu$V after baseline correction. The remaining pairs form
FIT/VAL/HEAD/EVAL sets of $622/288/716/1381$, and each EVAL participant
keeps both classes. Features are channel means over fifteen consecutive
50 ms bins covering 0--750 ms, i.e.\ 240 coordinates with no encoder or PCA;
this protocol is separate from the EEGPT-based reuse protocol below.

\textbf{Recorded SSVEP.}
Source 0/1 denotes actual dry/wet sessions, which lack physiological trial
correspondence. FIT/VAL/HEAD/EVAL contain $9840/4800/4800/5040$ observations
under one fixed stratified split. We construct six calibration views per
trial using spectral transport, spatial transport and their composition,
each at strengths .5 and 1 toward the opposite source. FIT prestimulus
signals determine smoothed log-power gains clipped to $[.25,4]$ and spatial
whitening/coloring with covariance shrinkage .1. MGPA corrects the token
mean and adds its displacement to every token; source readers and the
twelve-frequency task access the full $15\times512$ representation.

\subsection{Fitting, replay and comparison interfaces}
\label{app:fitting-replay}
\label{app:component-fitting}
The numerical settings use $d$ for the dimension of the representation being
corrected. The gradient stabilizer $\lambda$, damping coefficient $\tau$
and step radius $\rho$ control the iterative update, and $K$ is its maximum
number of stages, as in Algorithm~\ref{fig:algorithm-main}.
\begin{table}[tp]
\centering\small
\caption{\textbf{Shared MGPA settings.} Component ablations change only the
specified critic or target. LR denotes learning rate, $C$ is the logistic
regularization parameter, and $d$ is the corrected representation's
dimension. Split, reader and assignment seeds are separate.}
\label{tab:shared-settings}
\begin{tabular}{@{}>{\raggedright\arraybackslash}p{.34\linewidth}>{\raggedright\arraybackslash}p{.60\linewidth}@{}}
\toprule
Component & Setting \\
\midrule
Iterative fits / horizon & Seeds 17, 29, 43; at most 48 stages \\
Source critics & Balanced logistic ($C=1$); GELU MLP, hidden width 128 \\
Conditional-mean anchor & Ridge with intercept and penalty 10 (both constructions) \\
Neural-critic optimizer & AdamW, batch 256, weight decay $.001$ \\
Gradient stabilizer / damping & $.1$ / $.001$ \\
Controlled and real-representation fits & 12 critic epochs, LR $.001$, radius $.05\sqrt d$ \\
P300 reuse fits & At most 100 epochs, LR $.0003$, source-VAL patience 10,
radius $.025\sqrt d$ \\
Closed-form residual covariance & $.01$ shrinkage toward mean-diagonal identity \\
Participant uncertainty & 20,000 percentile bootstrap resamples, seed 1101 \\
\bottomrule
\end{tabular}
\end{table}

\begin{algorithm}[tp]
\small
\caption{Iterative MGPA: fit once, replay without source labels}
\label{fig:algorithm-main}
\begin{enumerate}
\setlength{\itemsep}{2pt}
\item \textbf{Inputs:} calibration pairs, source-labeled FIT/VAL embeddings,
gate rank or energy threshold, critic bank, $\lambda,\tau,\rho,K$, and
endpoint rule. Task-informed selection additionally requires DEV task data.
\item Fit standardization and the contrast-SVD gate (Section~\ref{sec:admissibility});
freeze $B,P,Q$. Return identity if $r=0$. Store original FIT/VAL complements.
\item Fit source critics on current FIT embeddings and regress their scores
on original FIT $Q$ to obtain conditional anchors.
\item Store critics and anchors; compute $e,A,W$, solve
\eqref{eq:tr-problem}, and apply $h\leftarrow h+B\Delta z_*$. Record VAL
outputs and repeat Steps 3--4 up to $K$ stages.
\item Select and store the prefix and optional final-stage fraction using
the declared validation rule; only task-informed selection uses task labels.
\item \textbf{Deployment:} standardize one embedding, replay the stored map,
and return it in the downstream interface's coordinates.
\end{enumerate}
\end{algorithm}

\textbf{Coordinates and calibration.}
Standardization uses FIT only, and gates use row-normalized contrasts.
Controlled and N170 fits filter contrast norms at the fifth percentile;
recorded SSVEP uses no percentile filter. Controlled gating uses at most
2048 deterministically sampled contrasts. Table~\ref{tab:protocol-map}
gives the retained-energy thresholds and ranks.

\textbf{Closed-form fitting.}
Closed-form MGPA shares the gate, standardization and Ridge anchor, and
uses the mean projected calibration contrast as its direction. Its metric
comes from residuals of a joint regression of $Z$ on $Q$ and source:
intercept/source coefficients are unpenalized and $Q$ has penalty 10.
The residual second moment is normalized by FIT count and shrunk as in
Table~\ref{tab:shared-settings}; \eqref{eq:closed-form-mgpa} uses linear
solves. Native studies take $\alpha=1$, while reuse selects $\alpha$ on
DEV. Recorded contrasts are oriented wet-like to dry-like for both the
gate and direction.

\textbf{Component ablations.}
The target ablation replaces the conditional anchor with the current-stage
empirical FIT-mean score; we also report a zero-score control. The
linear-only ablation removes the nonlinear critic. Other fitting settings
remain matched as specified with the corresponding results.

\textbf{Baselines and deployment information.}
LEACE fits an unrestricted covariance-weighted affine eraser to observed
FIT rows. CORAL uses FIT source means and identity-regularized sample
covariances, not pair correspondence. FEATMAP fits centered unregularized
least squares to FIT pairs, with the minimum-norm slope for rank-deficient
designs; controlled pairs use the evaluated $\gamma$-scaled endpoints.
CORAL/FEATMAP leave their chosen reference unchanged and require source
routing at deployment; both orientations are reported. MGPA, LEACE and
IGBP deploy source-blind. Recorded SSVEP has no paired FEATMAP comparison.
No method uses EVAL moments. Reuse searches regularization/strength on
P300 DEV instead of using these native settings.

\textbf{IGBP.}
\label{app:igbp-protocol}%
IGBP~\citep{iskander2023} shares each study's roles, standardized inputs and
independent readers. At each iteration a $d\to d\to2$ ReLU classifier
drives an ungated projective update without a conditional anchor or pair
correspondence. All reported fits use the longer-training preset: AdamW,
LR $2\times10^{-4}$, weight decay $.01$, batch 256, at most 50 epochs,
VAL source-accuracy patience 10 and improvement threshold $.002$, with no
$.80$ accuracy exit. Native maps apply 100 projections without task-based
stopping. Updates are evaluated directly in logit space to avoid softmax
saturation. For N170, $d=240$; for recorded SSVEP, $d=512$ and the resulting
mean correction is added to every token, matching MGPA's interface.

\textbf{Matched movement and edit-permission controls.}
\label{app:gate-attribution}%
Native outputs apply the complete fitted map, so their movements need not
match. A prescribed budget $b$ instead selects the first upward crossing
of $\|H'_{\rm VAL}-H_{\rm VAL}\|_F/\|H_{\rm VAL}\|_F=b$. Here
$H_{\rm VAL}$ and $H'_{\rm VAL}$ stack the original and corrected
validation representations, a prime denotes correction, and
$\|\cdot\|_F$ is the Frobenius norm. The scalar $b$ prescribes relative
movement. We interpolate only the last update and replay that prefix/fraction on HEAD/EVAL. The
N170 IGBP budget comparison uses the same first-crossing rule.

The controlled gate comparison uses the measurement subspace, leading
observed-FIT PCA directions of the same rank, five fixed random orientations
of that rank (averaged, not selected), and all 512 directions. Each arm
refits its critics and anchors with the same recipe and the original
measurement complement $Q_0$ as anchor input, where the subscript marks
the complement before any correction. Controls therefore receive
calibration-derived rank and conditioning information; only the measurement
arm necessarily preserves $Q_0$. Every arm and random orientation attains
the reported $.05$ budget, with all 108 cells, including identity, available.
At the $.10$ budget, 17 of 60 random trajectories do not reach it
within the horizon; we do not substitute their native endpoints.

\subsection{Independent readers and statistical summaries}
\label{app:independent-readouts}
\label{app:full-token-readers}
We train source readers on corrected HEAD outputs independently of the
adapters and evaluate them on participant-disjoint EVAL outputs.
Table~\ref{tab:reader-bank} lists their inputs and capacity. Reader
preprocessing uses HEAD only, with one example per recording rather than
per token. Neural readers use no early stopping and contribute three
restarts at 12/50/200 epochs.

\begin{table}[tp]
\centering\small
\caption{\textbf{Independent source-reader bank.} Controlled/N170 use the
11 common readers. Recorded SSVEP uses all $11+20+10=41$ readers; its common
bank alone defines $S_{\rm mean}$, the maximum source AUROC readable from
the pooled token mean. LR denotes learning rate and $C$ the logistic
regularization parameter. Arrows between network widths indicate
successive layer dimensions.}
\label{tab:reader-bank}
\begin{tabular}{@{}>{\raggedright\arraybackslash}p{.23\linewidth}>{\raggedright\arraybackslash}p{.71\linewidth}@{}}
\toprule
Input & Readers and training \\
\midrule
Common representation & Ordinary and covariance-whitened logistic ($C=1$,
whitening shrinkage $.01$); GELU MLP with width 128, AdamW, LR $.001$,
batch 256, weight decay $.001$. \\
\addlinespace
Direct full tokens & Raw logistic ($C=1$), HEAD-standardized logistic
($C=.01$); flat $7680\to128\to2$ GELU with common training; token network
with LayerNorm, shared $512\to64$ GELU, concatenation and binary output,
LR $.0005$, batch 128, weight decay $.001$, gradient clipping 5. \\
\addlinespace
Centered token residuals & Standardized logistic and the same token network
with its three restarts/checkpoints. \\
\bottomrule
\end{tabular}
\end{table}

For a recorded trial, let $H_t$ be token $t$'s feature vector and $\bar H$
the mean across its tokens. Common-offset correction preserves the centered
token residual $R_t=H_t-\bar H$. Residual readers fit once on original
HEAD residuals and are reused on corrected EVAL; numerical invariance is checked. This retained token
variation is distinct from the pooled gate complement $Q$.

\textbf{Task utility and movement.}
Controlled/N170 heads are HEAD-standardized balanced Ridge classifiers
(penalty 10, LSQR). The recorded task head uses HEAD-standardized flattened
features and the token architecture above with a twelve-class output,
trained for 50 epochs. Frozen heads and scalers train on original HEAD;
refitted heads and scalers train on corrected HEAD; both evaluate corrected
EVAL. The frozen and refitted scores $T_f/T_r$ equally average participant
AUROCs over $0\to1$ and $1\to0$ transfer, where each arrow points from
the head-training source to the evaluation source; recorded scores are
twelve-class macro AUROC. Movement $M$ averages
$\|H'_p-H_p\|_F/\|H_p\|_F$ over EVAL participants. Here $p$ indexes a
participant, and $H_p$ and $H'_p$ stack that participant's original and
corrected representations. This participant mean can differ from the
aggregate VAL movement used for endpoint selection.

\textbf{Aggregation and uncertainty.}
\label{app:evaluation-scope}%
For each fit, source accessibility is the maximum participant-mean AUROC
over readers; we then average across fits. Controlled scores pool disjoint
EVAL folds, with one output per split for deterministic maps. Shared
participant-bootstrap draws pair methods, transfer directions and fits,
recomputing the source maximum in every draw.

Intervals condition on fitted coordinates, adapters and readers. They are
pointwise and unadjusted for multiplicity; seeds and trials do not increase
participant counts. The controlled gate/target studies
share participants, and the two N170 protocols overlap. Source maxima
measure accessibility under the declared reader bank; they neither establish
complete erasure nor estimate unbiased performance of an EVAL-selected
reader.

\subsection{P300-selected correction reused on N170 and MMN}
\label{app:recovery-protocol}
\label{app:reuse-protocol}
\textbf{Calibration and shared coordinates.}
P300, N170 and MMN use simultaneous Neuroscan/Flex recordings and the
same eight EVAL participants. CAL contains the first 100 consecutive P300
pairs per FIT participant. We exclude any other FIT event whose one-second
epoch overlaps CAL on either device, leaving 2260 observed events. One
endpoint per observed event plus both endpoints of the 400 CAL pairs gives
3060 unique fitting endpoints for MGPA, LEACE and CORAL; FEATMAP uses the
400 pairs. CAL supplies the MGPA gates and Closed-form direction, while
Iterative critics fit the pooled bank without paired targets.

Frozen EEGPT retains $4\times512$ features per window, giving
$7\times4\times512=14{,}336$ features across seven overlapping windows,
with fixed $20\,\mu$V input scaling. Centered PCA256
retains 256 principal components; both PCA and standardization fit only
the Neuroscan endpoints of remaining P300 FIT events, then apply to both devices and all tasks. Neither N170 nor MMN
fits coordinates or selects the adapter. HEAD uses all technically valid
events, unlike the amplitude-filtered temporal N170 study.

\textbf{Frozen heads and association shifts.}
Each task has two balanced logistic heads trained on original HEAD features
with equal participant weights: an aligned, shortcut-sensitive head and an
independent-assignment control head. The logistic regularization parameter
is selected from $C\in\{10^{-6},10^{-5},\ldots,10\}$ by leave-one-HEAD-participant-out
independent-head AUROC and shared between that task's heads. Heads,
standardizers and assignment realizations stay fixed across adapter fits.
For participant $p$, let $\pi_p$ be positive-class prevalence and
$y\in\{0,1\}$ the event's task label. The parameter $\xi$ controls the
strength and direction of device--label association, and $U$ denotes the
selected device. Endpoint assignment satisfies
\[
 \Pr(U=1\mid Y=y,p;\xi)=\tfrac12+.4\xi
 \frac{y-\pi_p}{\max(\pi_p,1-\pi_p)},\qquad \xi\in[-1,1].
\]
Aligned/independent/reversed settings are $\xi=1/0/-1$; device marginals
remain one half and signals are unchanged. Fitting samples recorded
endpoints; evaluation weights both endpoints. Worst AUROC is the exact
minimum of the participant-mean quadratic AUROC curve on $[-1,1]$,
including interior minima. Control-head AUROC is evaluated at $\xi=0$.

\textbf{Selection and transfer.}
Both MGPA constructions use rank-eight gates and pooled conditional-mean
targets. Correction strength is the scalar $\alpha$ for the affine map or
the selected iterative prefix. Iterative training follows
Table~\ref{tab:shared-settings}, with 1332 DEV source-validation endpoints. P300 DEV selects the endpoint that
maximizes worst AUROC subject to at most $.01$ loss from identity for
\emph{each} head at independent association. Ties prefer less movement,
then an earlier endpoint.

All iterative prefixes are eligible. Affine strengths are
$0,.02,\ldots,1,1.1,1.2,1.3$; baseline ridge grids and both CORAL/FEATMAP
reference choices use the same selection rule. Across recipes within a
family, candidates within $.002$ of the best admissible DEV worst AUROC
are ordered by movement and the fixed tie rule. The repository's task-reuse
guide records the grids and selected values; Closed-form MGPA selects
$\alpha=1.1$. We then reuse each complete selected map unchanged on P300,
N170 and MMN.

Reuse movement for DEV tie-breaking is the participant mean of the ratio
of mean paired-event displacement norm to mean input norm, rather than the
Frobenius ratio above. Reported worst AUROC averages fit-specific minima
of participant-mean curves, not the minimum of an ensemble curve.
Each shared participant-bootstrap draw recomputes every curve and its
minimum before averaging fits; the same draws pair all tasks and methods.
Control-head AUROCs average participants and fits. Other uncertainty
conventions follow Appendix~\ref{app:evaluation-scope}.

\section{Supporting Experimental Results}
\label{app:retained-results}
This section reports paired contrasts, reader-level results and transfer
directions underlying the main comparisons. Unless a table specifies
percentage points, differences use the original AUROC scale. Bracketed
ranges are paired 95\% confidence intervals (CIs), computed as in
Appendix~\ref{app:independent-readouts}.

\subsection{Controlled correction and target choice}
\label{app:closed-form-controlled}
\label{app:target-controls}
\textbf{Closed-form correction.}
Closed-form MGPA reduces source accessibility with little observed task
change in controlled SSVEP. In Table~\ref{tab:controlled-cf}, its source
AUROC change from identity is $-.173$ $[-.180,-.164]$, while frozen- and
refitted-task changes are $-.0002$ $[-.0007,+.0002]$ and $+.0004$
$[-.0001,+.0008]$. The refitted-task gain over LEACE is $+.0879$
$[.0762,.0998]$. These contrasts support selective correction in this
constructed reference-change setting for both fixed and retrained heads.

\textbf{Conditional versus unconditional targets.}
The unconditional control targets each critic's empirical mean score on
current corrected FIT inputs. We recompute this mean at every stage
without downstream labels or HEAD/EVAL data, keeping the gate, critic
fitting, solver and 48-stage horizon unchanged. Each trajectory refits its
critics, so the comparison tests learned correction rather than the
fixed-score equivalence in Proposition~\ref{prop:parallel-levels}.
Zero-score targeting is a separate control because balanced source labels
do not imply a zero mean logit.

\begin{table}[tp]
\centering\small
\caption{\textbf{Conditional versus constant targets.}
Native controlled outputs for the same 81 participants and fits as
Figure~\ref{fig:target-cost}c. $S$: maximum source AUROC over the reader
bank; $T_f/T_r$: task AUROC with frozen/refitted heads; $M$:
participant-mean relative movement. FIT is the adapter-fitting set.
Arrows indicate the preferred direction.}
\label{tab:target-controls}
\begingroup\small\setlength{\tabcolsep}{4pt}
\begin{tabular}{@{}lrrrr@{}}
\toprule
Target & $S\downarrow$ & $T_f\uparrow$ & $T_r\uparrow$ & $M$ \\
\midrule
Conditional mean & 0.5105 & 0.7496 & 0.7527 & 0.1500 \\
Unconditional mean (FIT) & 0.5214 & 0.7488 & 0.7574 & 0.4063 \\
Zero score & 0.5199 & 0.7479 & 0.7562 & 0.4172 \\
\bottomrule
\end{tabular}\endgroup

\end{table}

Conditional anchoring lowers movement and source accessibility but gives
slightly lower refitted-task AUROC. Relative to empirical-unconditional
anchoring in Table~\ref{tab:target-controls}, the paired changes are
$-.2563$ $[-.2907,-.2232]$ in movement and $-.0110$ $[-.0143,-.0079]$ in
source AUROC. Frozen-task change is $+.0008$ $[-.0015,.0030]$, whereas
refitted-task change is $-.0047$ $[-.0070,-.0025]$. Thus the movement
advantage does not imply an advantage for every downstream readout.

\subsection{Temporal N170: nonlinear correction and task utility}
\label{app:critic-complexity}
\label{app:igbp-comparison}
\textbf{Linear-only ablation.}
We remove the GELU critic while retaining the gate, Ridge mean anchor,
smooth weighting, trust-region radius and stage horizon. Single-critic
damping $.0005$ preserves regularization relative to mean critic loss,
equivalently duplicating the affine constraint at the shared two-critic
damping. Because the bounded steps depend on the sample, this ablation
uses a linear critic but is not a globally affine map.

\begin{table}[tp]
\centering\small
\caption{\textbf{Nonlinear critics on N170.} A: $L$ and $S$ are maximum
source AUROCs over linear readers and the full reader bank, respectively,
for the ten participants in Table~\ref{tab:real-main}A.
B: $\Delta$ denotes Iterative minus linear-only, in AUROC percentage
points; $T_f/T_r$ denote frozen/refitted task AUROC. Brackets give paired
95\% confidence intervals (CIs), and arrows indicate the preferred
direction. Iterative estimates average fits; linear-only is deterministic.}
\label{tab:critic-complexity}
\label{tab:critic-complexity-paired}
\begingroup\small\setlength{\tabcolsep}{4pt}
\begin{tabular*}{\textwidth}{@{\extracolsep{\fill}}lrr@{}}
\toprule
\multicolumn{3}{@{}l}{\textbf{A. Native source accessibility}} \\
Correction & $L\downarrow$ & $S\downarrow$ \\
\midrule
Identity & 0.763 & 0.989 \\
MGPA-Iter & 0.794 & 0.946 \\
Linear-only & 0.663 & 0.987 \\
\bottomrule
\end{tabular*}\endgroup
\par\vspace{4pt}
\begingroup\small\setlength{\tabcolsep}{4pt}
\begin{tabular*}{\textwidth}{@{\extracolsep{\fill}}lrrr@{}}
\toprule
\multicolumn{4}{@{}l}{\textbf{B. MGPA-Iter minus Linear-only (AUROC percentage points)}} \\
Contrast & $\Delta S\downarrow$ [95\% CI] & $\Delta T_f\uparrow$ [95\% CI] & $\Delta T_r\uparrow$ [95\% CI] \\
\midrule
Paired difference & -4.13 [-6.47, -2.10] & +0.93 [-0.05, +1.88] & +1.71 [+0.05, +3.84] \\
\bottomrule
\end{tabular*}\endgroup

\end{table}

The nonlinear critic improves full-bank attenuation and refitted-task
utility, although linear accessibility does not follow the same pattern.
Table~\ref{tab:critic-complexity} shows that linear-only correction lowers
the strongest linear-reader score while leaving the full-bank score nearly
unchanged. Iterative MGPA lowers the full-bank score with higher task
scores at nearly the same movement, but its strongest linear-reader score
is higher than identity's. The ablation therefore supports an improvement
in the declared reader-bank maximum, not uniform attenuation across readers.

\textbf{Comparison with IGBP.}
Iterative MGPA improves N170 task utility over IGBP at native outputs and
at the larger matched movement budget. At native outputs, frozen/refitted
task gains are $+.0123$ $[.0042,.0203]$ and $+.0361$ $[.0306,.0405]$, with
movement lower by $.1504$ $[.1230,.1769]$; the source-AUROC difference is
$-.0091$ $[-.0198,.0006]$. At common VAL budget $.25$
(Table~\ref{tab:igbp-budget}), source, frozen-task and refitted-task
changes are $-.0234$ $[-.0393,-.0099]$, $+.0082$ $[.0015,.0154]$ and
$+.0156$ $[.0063,.0269]$. At budget $.10$, source accessibility is lower
but a downstream advantage is not established. The benefit thus depends
on the operating point; equal iteration counts would not equalize the
computation of these complete correction procedures.

\begin{table}[tp]
\centering\small
\caption{\textbf{N170: MGPA versus IGBP at matched VAL budgets.}
VAL/EVAL denote validation/evaluation participants. $S$ is the maximum
source-reader AUROC; $T_f/T_r$ are frozen/refitted task AUROC, and $M$ is
participant-mean relative movement. Arrows indicate the preferred
direction. First-crossing endpoints follow Appendix~\ref{app:gate-attribution};
attained EVAL movement $M$ may differ from the VAL budget. Same ten
participants and shared fits.}
\label{tab:igbp-budget}
\begingroup\small\setlength{\tabcolsep}{4pt}
\begin{tabular}{@{}llrrrr@{}}
\toprule
VAL budget & Method & $S\downarrow$ & $T_f\uparrow$ & $T_r\uparrow$ & $M$ \\
\midrule
.10 & MGPA-Iter & 0.9850 & 0.7274 & 0.7300 & 0.0974 \\
.10 & IGBP & 0.9893 & 0.7292 & 0.7284 & 0.0898 \\
\addlinespace[2pt]
.25 & MGPA-Iter & 0.9643 & 0.7292 & 0.7394 & 0.2422 \\
.25 & IGBP & 0.9877 & 0.7210 & 0.7237 & 0.2239 \\
\bottomrule
\end{tabular}\endgroup

\end{table}

\textbf{Task utility by transfer direction.}
Iterative MGPA's refitted-task gain over identity is concentrated in
Neuroscan-to-Flex transfer. The frozen/refitted gains are $+.0070$
$[+.0012,+.0131]$ and $+.0187$ $[+.0042,+.0366]$, respectively, with equal
weight on both directions. Refitted AUROC rises from $.608$ to $.648$
for Neuroscan-to-Flex, while the reverse direction changes from $.847$
to $.844$. The refitted-task improvement is therefore specific to one transfer
direction.

\subsection{Recorded SSVEP: full-input accessibility and utility}
\label{app:recorded-full-input}
MGPA's source attenuation extends from pooled features to readers of the
complete token representation. On the same saved outputs and participant
roles, Iterative MGPA gives pooled source AUROC $.7954$ and full-bank
$S=.8718$, versus identity's $.9155$. The maximum among readers trained
directly on all tokens falls from $.8987$ to $.8685$, a paired change of
$-.0302$ $[-.0459,-.0174]$; the full-bank difference from LEACE is
$-.0463$ $[-.0671,-.0225]$. Closed-form MGPA gives $S=.8949$, a change of
$-.0206$ $[-.0356,-.0043]$ from identity. All maps share the 41-reader bank
(Appendix~\ref{app:full-token-readers}), so the reduction is not limited
to auditing the corrected token mean.

Unchanged token residuals set a shared source-accessibility floor for
several methods. Residual-only readers attain $.8718$ for every
common-offset correction and determine the full-bank maximum for Iterative
MGPA, linear-only MGPA and IGBP. This maximum therefore includes retained
temporal structure that the shared token displacement cannot change.
CORAL reaches $S=1$ in both orientations, because its per-source offset
is itself a source signature under this common-offset interface;
token-specific CORAL maps are not evaluated.

\textbf{Comparison with IGBP and task utility.}
IGBP matches Iterative MGPA's full-bank source score with less movement
and similar task performance. Both attain $S=.8718$, with pooled-mean
scores $.7832$ and $.7954$, respectively. IGBP uses movement $.0956$
versus $.1490$, has slightly higher frozen-task AUROC ($.6891$ versus
$.6880$), and similar refitted-task AUROC ($.6884$ versus $.6886$).
Against identity, Iterative MGPA's frozen/refitted task changes are
$-.0002$ $[-.0017,+.0013]$ and $+.0004$ $[-.0028,+.0034]$. These results
support little observed task change after attenuation, without establishing
an Iterative MGPA advantage in this setting.

\subsection{P300-selected reuse: absolute outcomes and paired comparisons}
\label{app:transfer-results}
\label{app:reuse-results}
Both MGPA corrections improve transferred worst-association AUROC over
LEACE on N170 and MMN at the unchanged P300-selected operating points.
Tables~\ref{tab:reuse-absolute}--\ref{tab:reuse-paired} show paired intervals
excluding zero on both transfer tasks, while the P300 intervals contain
zero. Control-head changes from identity are small: Closed-form MGPA gains
$.0012$ $[.0006,.0018]$ on MMN, and the other three MGPA transfer-task
intervals contain zero. N170 worst AUROC remains below .5, so the
improvement is partial robustness recovery under the induced shift.

\begin{figure}[tp]
\centering
\includegraphics[width=\linewidth]{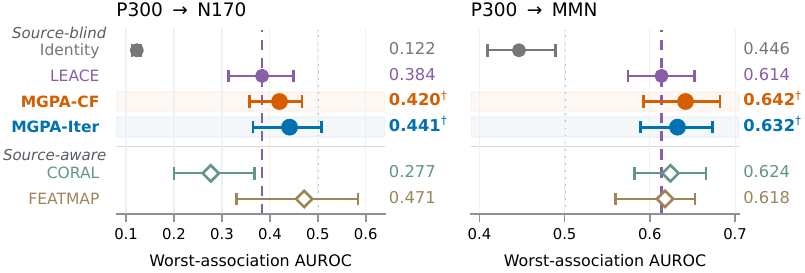}
\caption{\textbf{Cross-task robustness of a P300-selected correction.}
Absolute worst-association AUROC of frozen N170/MMN shortcut-trained heads;
whiskers show marginal 95\% participant-bootstrap intervals.
The horizontal axis reports AUROC, with larger values indicating better
worst-association performance. Dashed: LEACE; dotted: chance;
hollow: source-aware methods.
$\dagger$: paired 95\% interval for the gain over LEACE entirely above zero.
Intervals condition on fitted maps and are unadjusted for multiplicity;
panel ranges differ.}
\label{fig:reuse}
\end{figure}

\begin{table}[tp]
\centering\small
\caption{\textbf{Absolute reuse outcomes.} A/I/R: aligned/independent/reversed
AUROC of the shortcut-sensitive head. These regimes respectively preserve,
remove and reverse the device--label association used to train that head;
worst/control results are in Table~\ref{tab:reuse-main}. Iterative entries average fits; other maps are
evaluated once. CORAL/FEATMAP require device identity.}
\label{tab:reuse-absolute}
\begingroup\small\setlength{\tabcolsep}{4pt}
\begin{tabular*}{\textwidth}{@{\extracolsep{\fill}}lrrrrrrrrr@{}}
\toprule
 & \multicolumn{3}{c}{P300} & \multicolumn{3}{c}{N170} & \multicolumn{3}{c}{MMN} \\
\cmidrule(lr){2-4}\cmidrule(lr){5-7}\cmidrule(lr){8-10}
Method & A & I & R & A & I & R & A & I & R \\
\midrule
Identity & 0.770 & 0.568 & 0.355 & 0.914 & 0.547 & 0.122 & 0.753 & 0.603 & 0.446 \\
LEACE & 0.604 & 0.602 & 0.601 & 0.782 & 0.590 & 0.384 & 0.614 & 0.633 & 0.652 \\
MGPA-CF & 0.643 & 0.621 & 0.599 & 0.735 & 0.579 & 0.420 & 0.642 & 0.644 & 0.646 \\
MGPA-Iter & 0.636 & 0.617 & 0.599 & 0.744 & 0.594 & 0.441 & 0.632 & 0.639 & 0.646 \\
CORAL & 0.662 & 0.607 & 0.550 & 0.818 & 0.564 & 0.277 & 0.639 & 0.632 & 0.624 \\
FEATMAP & 0.625 & 0.600 & 0.573 & 0.697 & 0.591 & 0.471 & 0.623 & 0.621 & 0.618 \\
\bottomrule
\end{tabular*}\endgroup

\end{table}

\begin{table}[tp]
\centering\small
\caption{\textbf{Paired reuse gains.} $\Delta$ is the named method's AUROC
minus that of the reference method, on the original AUROC scale; positive
values indicate improvement. ``Worst'' refers to worst-association AUROC
of the shortcut-sensitive head; ``control head'' uses independent device
assignment. Brackets give paired 95\% confidence intervals (CIs), with
worst AUROC recomputed in each bootstrap draw
(Appendix~\ref{app:reuse-protocol}).}
\label{tab:reuse-paired}
\begingroup\small\setlength{\tabcolsep}{4pt}
\begin{tabular*}{\textwidth}{@{\extracolsep{\fill}}lllrr@{}}
\toprule
Task & Method & Reference & $\Delta$ worst [95\% CI] & $\Delta$ control head [95\% CI] \\
\midrule
P300 & MGPA-CF & Identity & $+0.244\ [+0.220, +0.268]$ & $+0.0005\ [-0.0010, +0.0022]$ \\
P300 & MGPA-CF & LEACE & $-0.002\ [-0.022, +0.027]$ & $+0.0050\ [+0.0036, +0.0066]$ \\
P300 & MGPA-Iter & Identity & $+0.244\ [+0.224, +0.262]$ & $+0.0003\ [-0.0016, +0.0025]$ \\
P300 & MGPA-Iter & LEACE & $-0.002\ [-0.016, +0.021]$ & $+0.0047\ [+0.0034, +0.0062]$ \\
N170 & MGPA-CF & Identity & $+0.298\ [+0.244, +0.340]$ & $-0.0014\ [-0.0044, +0.0019]$ \\
N170 & MGPA-CF & LEACE & $+0.036\ [+0.005, +0.067]$ & $+0.0002\ [-0.0023, +0.0037]$ \\
N170 & MGPA-Iter & Identity & $+0.319\ [+0.251, +0.380]$ & $-0.0010\ [-0.0035, +0.0013]$ \\
N170 & MGPA-Iter & LEACE & $+0.057\ [+0.034, +0.083]$ & $+0.0006\ [-0.0009, +0.0025]$ \\
MMN & MGPA-CF & Identity & $+0.195\ [+0.168, +0.215]$ & $+0.0012\ [+0.0006, +0.0018]$ \\
MMN & MGPA-CF & LEACE & $+0.028\ [+0.002, +0.034]$ & $+0.0024\ [+0.0006, +0.0043]$ \\
MMN & MGPA-Iter & Identity & $+0.186\ [+0.157, +0.209]$ & $+0.0003\ [-0.0014, +0.0017]$ \\
MMN & MGPA-Iter & LEACE & $+0.019\ [+0.004, +0.024]$ & $+0.0016\ [-0.0005, +0.0036]$ \\
\bottomrule
\end{tabular*}\endgroup

\end{table}

The transferred worst-association gain over LEACE holds in each Iterative
MGPA fit. Across the three fits, worst AUROC ranges from $.417$ to $.471$
on N170 and $.622$ to $.639$ on MMN; every P300-selected map improves this
score when reused unchanged. Table~\ref{tab:reuse-paired} reports intervals
for the mean across fits, while these ranges describe variation among the
three fitted maps.
\FloatBarrier

\end{document}